\documentclass[]{template}

\usepackage{xcolor}
\usepackage[utf8]{inputenc} 
\usepackage{hyperref}       
\definecolor{airforceblue}{rgb}{0.36, 0.54, 0.66}
\definecolor{bluegray}{rgb}{0.4, 0.6, 0.8}
\definecolor{bleudefrance}{rgb}{0.19, 0.55, 0.91}
\hypersetup{colorlinks,linkcolor={airforceblue},citecolor={bleudefrance},urlcolor={bluegray}}
\usepackage{url}            
\usepackage{booktabs}       
\usepackage{amsfonts}       
\usepackage{nicefrac}       
\usepackage{microtype}      
\usepackage{pifont}
\usepackage{fontawesome5}
\usepackage{natbib}

\usepackage{graphicx}
\usepackage{multirow}
\usepackage{array}
\usepackage{colortbl}
\usepackage{makecell}
\definecolor{rlblue}{RGB}{214,234,248}
\definecolor{sftgray}{RGB}{235,237,239}
\usepackage{amsmath}
\usepackage{caption}
\usepackage[most]{tcolorbox}
\usepackage{needspace}
\usepackage[table]{xcolor}
\usepackage{listings}

\usepackage{tabularx} 
\usepackage{array} 
\usepackage{siunitx}

\newcolumntype{L}{>{\raggedright\arraybackslash}X} 
\newcommand{\avgcell}[1]{\cellcolor{blue!8}{#1}}

\newcolumntype{L}[1]{>{\raggedright\arraybackslash}p{#1}}
\newcolumntype{C}[1]{>{\centering\arraybackslash}p{#1}}
\newcommand{\cmark}{\textcolor{green!60!black}{\ding{51}}}
\newcommand{\xmark}{\textcolor{red}{\ding{55}}}
\newcommand{\eg}{\textit{e.g.}}
\newcommand{\ie}{\textit{i.e.}}
\newcommand{\sysrl}{OpenWebRL-4B}
\newcommand{\syssft}{OpenWebRL-4B-SFT}

\title{OpenWebRL: Demystifying Online Multi-turn Reinforcement Learning for Visual Web Agents}

\author[]{%
Rui Yang$^{1*\dagger}$,\hspace{0.4em}
Qianhui Wu$^{2*\ddagger}$,\hspace{0.4em}
Yuxi Chen$^{1}$,\hspace{0.4em}
Hao Bai$^{1}$,\hspace{0.4em}
Wenlin Yao$^{2}$,\hspace{0.4em}
Hao Cheng$^{2}$  \\
Baolin Peng$^{2}$,\hspace{0.4em} 
Huan Zhang$^{1}$,\hspace{0.4em}
Tong Zhang$^{1}$,\hspace{0.4em}
Jianfeng Gao$^{2}$ %

}

\affiliation{
\includegraphics[height=4mm, trim=100 55 100 55, clip]{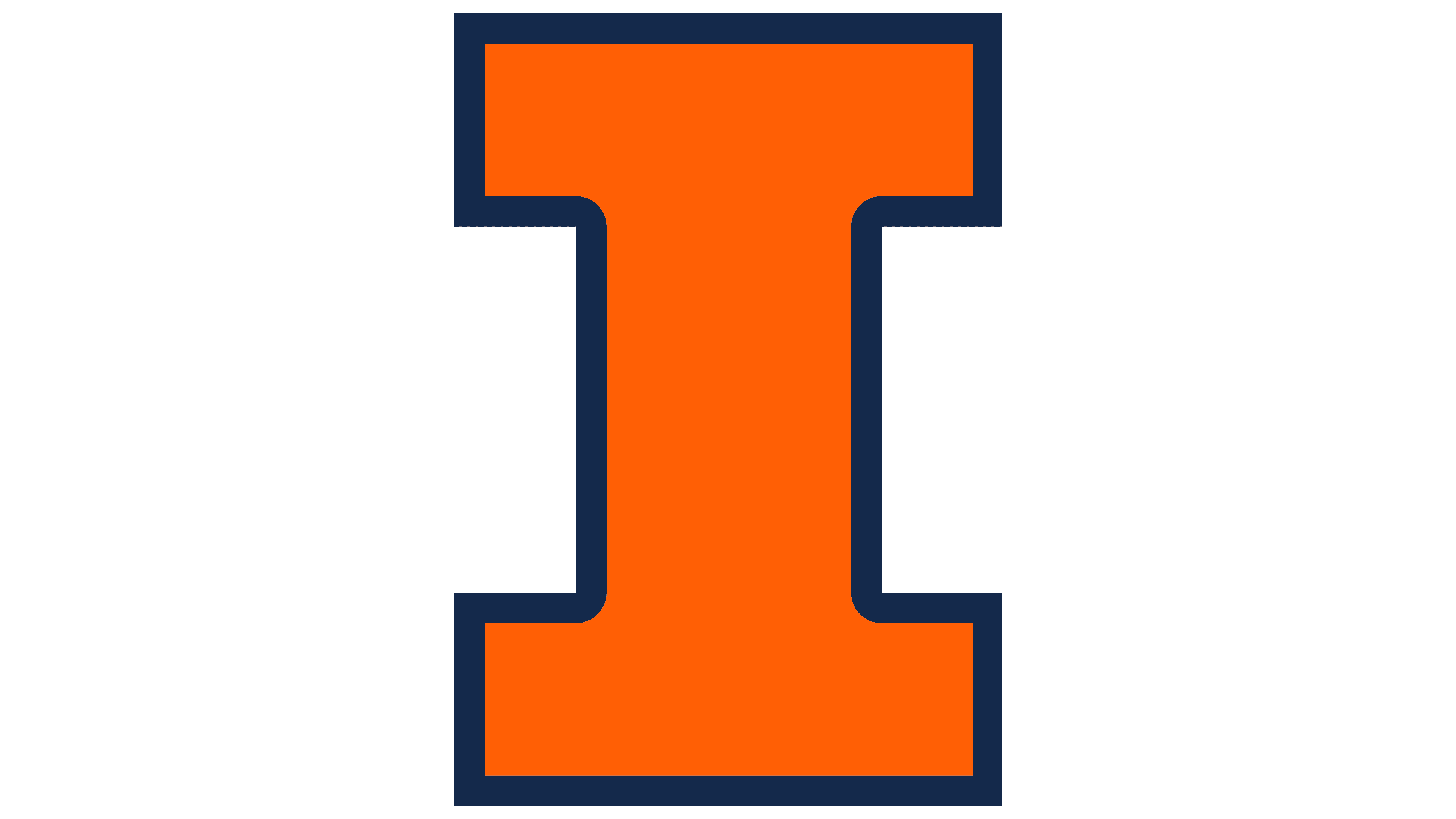} \hspace{-7pt} $^1$UIUC, \hspace{5pt}
\includegraphics[height=4mm, trim=55 55 55 55, clip]{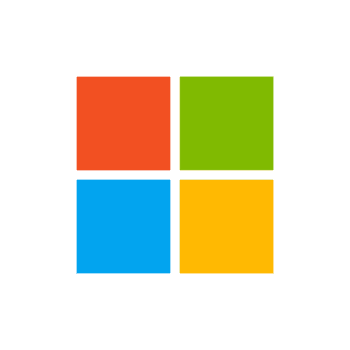} $^2$Microsoft
\\
\vspace{6pt}
\href{https://openwebrl.github.io}{\faGlobe \ https://openwebrl.github.io}
}

\usepackage{xcolor}
\usepackage{makecell}
\usepackage{url}
\usepackage{breakurl}

\tcbset{
    promptboxstyle/.style={
        colback=blue!5!white,
        colframe=blue!75!black,
        fonttitle=\bfseries,
        enhanced jigsaw,
        breakable,
        parbox=false,
        boxrule=0.8pt,
        left=6pt,
        right=6pt,
        top=6pt,
        bottom=6pt,
        pad at break*=1mm
    }
}

\newtcolorbox{promptbox}[2][]{
    colback=blue!5!white,
    colframe=blue!75!black,
    fonttitle=\bfseries,
    title={#2},
    enhanced jigsaw,
    breakable,
    parbox=false,
    boxrule=0.8pt,
    left=6pt,
    right=6pt,
    top=6pt,
    bottom=6pt,
    before={\par\medskip\Needspace{10\baselineskip}},
    after={\par\medskip},
    pad at break*=1mm,
    #1
}

\newcommand{\promptboximage}[3][0.31\linewidth]{%
    \begin{minipage}[t]{#1}
        \centering
        \includegraphics[width=\linewidth]{#2}\\[2pt]
        {\small #3}
    \end{minipage}
}

\begin{document}

\paperabstract{
Building capable visual web agents requires long-horizon reasoning, precise grounding, and robust interaction with dynamic real-world websites. Despite rapid progress, the strongest systems remain largely proprietary, while open agents still depend heavily on supervised post-training over large collections of curated web trajectories.
This dependence creates a major scalability bottleneck: high-quality demonstrations are expensive to collect, and static datasets offer limited coverage of the diverse, ever-changing open web. Although online reinforcement learning (RL) has shown promise for text-based agents, its potential for training visual web agents directly on live websites remains largely underexplored.
In this paper, we introduce \textbf{OpenWebRL}, an open framework for training visual web agents with online multi-turn RL on real websites. OpenWebRL covers the full training pipeline, including scalable live-browser infrastructure, supervised initialization, multimodal context management, trajectory-level success judging, and efficient multi-turn policy optimization. Using this framework, we train \textbf{OpenWebRL-4B}, which establishes a new open-source state of the art on challenging live-web benchmarks. With only 0.4K initialization trajectories and 2.2K open-ended RL training tasks, OpenWebRL-4B achieves 67.0\% success on Online-Mind2Web and 64.0\% on DeepShop, outperforming prior open agents of similar or larger scale and remaining competitive with proprietary systems including OpenAI CUA and Gemini CUA.
Beyond strong benchmark performance, we systematically study the key design choices that make online RL effective for visual web agents, and analyze how RL improves agentic reasoning. Overall, our work offers a practical path toward building more capable, reproducible, and cost-efficient open web agents. We will release our training data, models, and code to support future research.
}

\maketitle

\begin{figure}[h]
    \centering
    \includegraphics[width=1\linewidth]{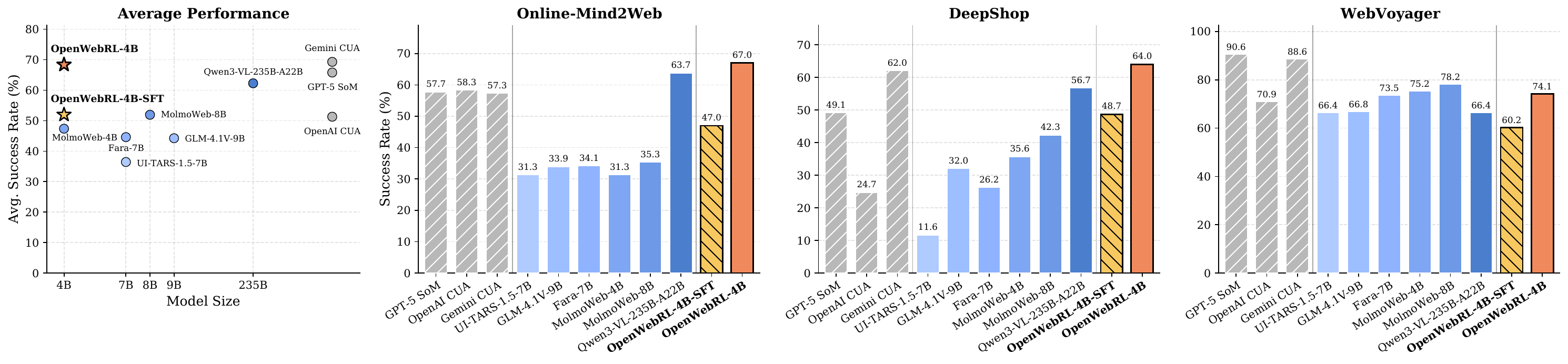}
    \vspace{-15pt}
\caption{Performance comparison on online web benchmarks, including Online-Mind2Web (2025.04)~\cite{xue2025an}, DeepShop (2025.06)~\cite{lyu2025deepshop}, and WebVoyager (2024.01)~\cite{he2024webvoyager}. 
As WebVoyager was constructed earlier than the other live benchmarks, temporal changes in websites may have a larger effect on direct comparability with previously reported results.
}
    \label{fig:result_illustration}
\end{figure}

\section{Introduction}

Web agents powered by large vision-language models (VLMs)~\cite{achiam2023gpt,bai2025qwen3,team2025kimi,singh2025openai,wang2025uitars2} have made rapid progress on realistic browser tasks, including product search, information extraction, and multi-step navigation~\cite{deng2023mind2web,he2024webvoyager,he2025openwebvoyager,xue2025an,zheng2025deepresearcher,xu2024aguvis,gou2024navigating,liu2025scalecua}. However, the strongest open agents still lag behind proprietary systems in long-horizon planning and adaptation to dynamic websites. At the same time, proprietary models, closed training data and pipelines, and expensive API-based evaluation create substantial barriers for open research. As a result, there remains a widening gap between what closed systems can achieve and what the open research community can study, reproduce, and build upon.

A central bottleneck is the lack of scalable data and training frameworks. Leading open agents such as MolmoWeb~\cite{gupta2026molmoweb} rely on supervised post-training over hundreds of thousands of curated trajectories, e.g., 278K trajectories, which are expensive to collect and inherently limited by static coverage~\cite{fara7b2025,gupta2026molmoweb,he2025webstar}. Online reinforcement learning offers a principled alternative: instead of imitating fixed demonstrations, agents can improve by interacting with real websites and learning from task outcomes. Yet applying online RL to visual agents on the open web introduces challenges that are qualitatively different from those in controlled environments: web pages are dynamic and non-stationary, live browser interaction is slow and brittle, and open-ended tasks often lack reliable rule-based verification. Prior RL work has largely sidestepped these challenges by focusing on text-only agents in simulated or self-hosted environments~\cite{wang2025ragen,zhang2025agentrl,wei2025webagent}. The closest efforts on live websites, such as PAE~\cite{zhou2024pae} and WebGym~\cite{bai2026webgym}, remain limited in their RL formulation and still rely on costly proprietary evaluation. More broadly, the design choices that make online RL effective for visual web agents remain underexplored.

We introduce \textbf{OpenWebRL}, a fully open framework for training visual web agents through direct interaction with live websites. Rather than treating online RL as a black-box recipe, OpenWebRL exposes and systematically studies the key factors behind effective open-web learning. The framework is built on a robust browser infrastructure for large-scale parallel rollout collection~\cite{peng2026orchard}, and consists of three components: (1) a \textit{supervised warm start} using only 0.4K trajectories, placing the policy in a productive exploration regime before online training; (2) an \textit{agent harness} with multi-tool action execution, textual environment feedback, and multimodal context management to make online agentic RL more efficient and reliable; and (3) a \textit{multimodal multi-turn GRPO objective} with trajectory-level judging, using either a GPT-based or a distilled 8B judge that matches proprietary performance while reducing evaluation cost by roughly \$545.5 per experiment.
Built with this framework, \textbf{OpenWebRL-4B} establishes a new state of the art among open visual web agents across three challenging live web benchmarks. With only a 4B backbone, 0.4K warm-start trajectories, and 2.2K open-web RL training tasks, OpenWebRL-4B achieves success rates of 74.1\% on WebVoyager~\cite{he2024webvoyager}, 67.0\% on Online-Mind2Web~\cite{xue2025an}, and 64.0\% on DeepShop~\cite{lyu2025deepshop}. These results substantially outperform prior open agents, including FARA-7B~\cite{fara7b2025}, MolmoWeb-8B~\cite{gupta2026molmoweb}, and even Qwen3-VL-235B-A22B-Thinking on the evaluated benchmarks, while remaining competitive with several proprietary systems such as GPT-5, OpenAI CUA, and Gemini CUA. Beyond benchmark performance, we systematically analyze the key design choices that make online RL effective for visual web agents and study how RL shapes agentic reasoning during training.

Our contributions are threefold. First, we introduce OpenWebRL, a fully open framework for end-to-end online RL of visual web agents on live websites. Second, we develop a practical multimodal multi-turn RL recipe that combines robust browser infrastructure, trajectory-level judging, and efficient context management, making open-web training effective for compact models. Third, we release a strong 4B-scale open agent together with a detailed empirical study of the ingredients that make online RL successful for visual web interaction. We hope OpenWebRL provides a practical foundation for future research on capable, reproducible, and cost-efficient open web agents.

\begin{table}[th]
\centering
\caption{%
  Summary of existing multi-turn web agents training pipelines.
  Simulated web refers to environments such as WebShop; self-hosted web refers to more realistic offline web stacks such as WebArena.
  Green \cmark\ / red \xmark\ indicate whether weights, data, and code were publicly released.
}
\label{tab:web-agents}
\scriptsize
\setlength{\tabcolsep}{3pt}
\renewcommand{\arraystretch}{1.25}
\resizebox{1\textwidth}{!}{%
\begin{tabular}{%
  L{2.5cm}   
  L{3cm}   
  L{1.8cm}   
  L{2.9cm}   
  L{2.0cm}   
  C{1cm}   
  C{1cm}   
  C{1cm}   
}
\toprule
\textbf{Paper} & \textbf{Method} & \textbf{Observation} & \textbf{Train Env} & \textbf{Reward} & \textbf{Weights} & \textbf{Data} & \textbf{Code} \\
\midrule


WebRL~\cite{qi2024webrl}
  & Online curriculum RL
  & Text
  & Self-hosted web
  & Trained judge
  & \cmark & \cmark & \cmark \\

AgentRL~\cite{zhang2025agentrl}
  & SFT+GRPO
  & Text
  & Simulated web
  & Rule-based
  & \xmark & \cmark & \cmark \\

PAE~\cite{zhou2024pae}
  & Online Filtered BC
  & Multimodal
  & Open web
  & Prompted judge
  & \cmark & \cmark & \cmark \\

RAGEN~\cite{wang2025ragen}
  & StarPO
  & Text
  & Simulated web
  & Rule-based
  & \xmark & \cmark & \cmark \\

WebAgent-R1~\cite{wei2025webagent}
  & SFT+Multi-turn GRPO 
  & Text
  & Self-hosted web
  & Rule-based
  & \xmark & \cmark & \cmark \\

UI-TARS-2~\cite{wang2025uitars2}
  & Multi-turn PPO
  & Multimodal
  & Computer-use sandbox
  & Mixed judges
  & \xmark & \xmark & \xmark \\

WebGym~\cite{bai2026webgym}
  & Online Filtered BC
  & Multimodal
  & Open web
  & Prompted judge
  & \xmark & \cmark & \cmark \\

\midrule


WebSTAR~\cite{he2025webstar}
  & Step-filtered SFT
  & Multimodal
  & ---
  & Prompted judge
  & \xmark & \cmark & \cmark \\

GUI-Libra~\cite{yang2026guilibra}
  & Step-wise GRPO
  & Multimodal
  & ---
  & Rule-based
  & \cmark & \cmark & \cmark \\

ScaleCUA~\cite{liu2025scalecua} & SFT & Multimodal & --- & ---  & \cmark & \cmark & \cmark \\

Fara-7B~\cite{fara7b2025}
  & SFT
  & Multimodal
  & ---
  & ---
  & \cmark & \xmark & \xmark \\

MolmoWeb~\cite{gupta2026molmoweb}
  & SFT
  & Multimodal
  & ---
  & ---
  & \cmark & \cmark & \cmark \\

\midrule
OpenWebRL (Ours)
  & SFT + MM-GRPO
  & Multimodal
  & Open web 
  & Mixed judges
  & \cmark & \cmark & \cmark \\

\bottomrule
\end{tabular}
}
\end{table}

\section{Related Work}
\textbf{LLM/VLM-based Web Agents.}
Recent advances in LLM/VLM-based web agents have been driven by three complementary threads: 
stronger foundation models~\cite{bai2025qwen3,team2025kimi,chen2024internvl,yang2025magma,hong2025glm}, agent frameworks~\cite{gu2024your,yang2025agentoccam,wang2026webxskill}, 
and web- or GUI-specific post-training~\cite{xu2024aguvis,qin2025ui,cheng2024seeclick,
wu2024atlas,wu2025gui,wang2025uitars2,gou2024navigating,chae2025web,zhuang2026workforceagent}. More recent systems further enhance agent 
capabilities through curated demonstrations, synthetic data, and supervised fine-tuning 
tailored to realistic browser interaction~\cite{liu2025scalecua,fara7b2025,
gupta2026molmoweb,he2025webstar}. In parallel, evaluation has broadened from curated, 
static web benchmarks toward more diverse and realistic open-web 
settings~\cite{pan2024webcanvas,he2024webvoyager,xue2025an,lyu2025deepshop,
trabucco2025insta,hong2026embodied,yu2026visual,chen2026captcha,bai2026webgym}.

\textbf{Reinforcement Learning for Multimodal Agents.}
Deploying VLMs as agents in visually grounded environments requires moving beyond static understanding toward long-horizon decision-making\cite{yang2025embodiedbench,wang2026vagen}. To bridge the gap, recent research adopts a two-stage SFT-then-RL paradigm\citep{chen2025era,zhai2024finetuninglargevisionlanguagemodels,zhan2025visual}.
The success of outcome-based RL for language 
reasoning~\cite{guo2025deepseek} has spurred growing efforts to extend RLVR to VLMs 
and interactive agents. Early work applies this paradigm to static tasks such as 
perception, grounding, and visual question 
answering~\cite{liu2025visual,huang2025vision,shen2025vlm,wang2025vl,lu2026ui,
luo2025gui,yang2026guilibra}, while more recent methods target multi-turn and agentic 
settings~\cite{qi2024webrl,
zhou2024pae,bai2024digirl,bai2026webgym,wang2025ragen,zhang2025agentrl,
wei2025webagent,wang2025uitars2}. However, most of this work operates in simulated, 
self-hosted, or narrowly scoped environments. In contrast, we study fully open online 
RL for compact visual web agents trained and evaluated on live websites.

\section{Preliminaries}
\textbf{Problem Formulation.} We formulate multimodal web-agent training as a POMDP 
$\mathcal{M} = (\mathcal{S}, \mathcal{O}, \mathcal{A}, \mathcal{T}, \mathcal{R})$. 
Each task $q$ specifies a start URL and instruction. At step $t$, the agent observes 
$o_t = (x_t, I_t)$, where $x_t$ contains textual browser information (URL, tab info, 
environment feedback) and $I_t$ is a screenshot. Given interaction history 
$h_t = (q, o_0, a_0, \ldots, a_{t-1}, o_t)$, policy $\pi_\theta$ generates a response 
$y_t$ containing reasoning and a structured browser action, after which the environment transitions to $s_{t+1} \sim \mathcal{T}(s_{t+1} \mid s_t, a_t)$ and yields $o_{t+1}$. 
An episode $\tau = \{(h_t, y_t, a_t, o_{t+1})\}_{t=0}^{T-1}$ terminates when the agent 
calls \texttt{done}, the step budget is exhausted, or an environment failure occurs. 
Task success is evaluated only after the full interaction via rule-based checks or a judge model.

\textbf{Multi-turn GRPO.}
GRPO~\cite{shao2024deepseekmath} is a critic-free policy optimization method that replaces a learned value function with group-relative advantages. 
In multi-turn settings, the sampled unit is a full trajectory rather than a single response. For each task $q$, the policy samples a group of trajectories $\{\tau_i\}_{i=1}^{G}$, where each trajectory receives a trajectory-level reward $R(\tau_i)$. The group-relative advantage is $
A_i = \frac{R(\tau_i) - \mathrm{mean}(\{R(\tau_j)\}_{j=1}^{G})}{\mathrm{std}(\{R(\tau_j)\}_{j=1}^{G})}$.
Multi-turn GRPO~\cite{wei2025webagent} propagates this trajectory-level signal to the action tokens generated across all turns in $\tau_i$. Denoting by $y_{i,t,k}$ the $k$-th token in the response at turn $t$ of trajectory $\tau_i$, the optimization objective becomes
\[
\mathcal{L}_{\mathrm{MT\text{-}GRPO}}(\theta)
=
-
\frac{1}{G}
\sum_{i=1}^{G}
\frac{1}{T_i}
\sum_{t=0}^{T_i-1}
\frac{1}{|y_{i,t}|}
\sum_{k=1}^{|y_{i,t}|}
\min\!\left(
\rho_{i,t,k}(\theta) A_i,\;
\operatorname{clip}\!\left(\rho_{i,t,k}(\theta), 1-\epsilon, 1+\epsilon\right) A_i
\right),
\]
where $\rho_{i,t,k}(\theta)$ is the token-level importance ratio at turn $t$.

\section{OpenWebRL: An Online RL Training Framework for Visual Web Agents}
\label{sec:openwebrl}
We train multimodal web agents with an end-to-end online RL pipeline on the open
web, starting from a general-purpose VLM. Our goal is not only to improve visual
web-agent performance, but also to systematically study the key components of
online RL for open-web navigation. 
Since online RL is computationally expensive, we focus our main experiments on
small VLMs, such as Qwen3-VL-4B \citep{bai2025qwen3}. However, general-purpose small VLMs often lack the web-specific knowledge needed to operate in dynamic browser environments. We therefore warm-start the policy with supervised finetuning on a small set of trajectories collected using a stronger open-source model.
Starting from the SFT policy, we iteratively collect online rollouts in live
browser environments and update the model with group-relative policy
optimization. In the following subsections, we describe the data preparation, rollout system, reward design, and optimization objective.

\subsection{Agent Harness}
\label{sec:harness}
Training and evaluating web agents on live websites introduces substantial
environment noise, including dynamic page updates, pop-ups, redirects, bot
detection, blocking, and transient network failures. To make open-web rollouts
more reliable, we build a fault-tolerant browser environment based on Orchard Env~\citep{peng2026orchard} with navigation retries, timeout handling, and structured failure attribution, separating unstable website behavior from model
behavior and making failures diagnosable during large-scale training. For efficiency, we run parallel browser instances that asynchronously sample independent trajectories, each maintaining its own page state and interaction history. Implementation details are in Appendix~\ref{app:env_infra}.

Based on the environment infrastructure, we employ a generic multi-turn ReAct-style tool-calling agent~\citep{yao2023react} to avoid conflating scaffold-specific effects with differences in data or training recipes, and more importantly, let the same agent paradigm generalize beyond GUI navigation to any tool-using domain. Within this framework, the only browser-specific design choices are how observations are organized, what tools the agent can call, and how multi-step interaction history is maintained.

\textbf{Observation and Environment Feedback.}
The environment provides a multimodal observation consisting of the current screenshot, active URL, viewport dimensions, and tab metadata. These signals allow the agent to track browser state across heterogeneous websites with diverse layouts, dynamic content, and multi-tab workflows.
In addition, \textbf{\emph{each action returns a concise textual environment-feedback message extracted from DOM-tree changes}} between consecutive interaction steps. These messages summarize the execution outcome and observable state changes, such as successful page navigation, new-tab creation, typed-text mismatches, or failed scrolling attempts. This lightweight feedback makes web interaction more observable, allowing the agent to distinguish successful actions from silent failures or unexpected browser behavior commonly encountered on live websites. More details are provided in Appendix~\ref{app:env_feedback}.

\textbf{Action Space and Multi-tool-call Interface.}
The agent is equipped with a structured action space $\mathcal{A}$ of $13$ atomic browser tools spanning 
pointer management (\texttt{click}, \texttt{hover}, \texttt{drag}), keyboard input (\texttt{write}, \texttt{press\_keys}), page navigation (\texttt{scroll}, \texttt{goto\_url}, \texttt{go\_back}, \texttt{wait}), tab management (\texttt{new\_tab}, \texttt{switch\_tab}, \texttt{close\_tab}), and termination via \texttt{done(response)},
which is the sole mechanism for successfully ending an episode and emitting the final user-facing answer.
At each step, the policy response is expected to consist of one reasoning block followed by \textbf{\textit{one or more tool call blocks}}. 
The environment parses the tool calls 
and executes the corresponding browser actions sequentially before returning per-call feedback and the next screenshot. This \textbf{\emph{multi-tool-call interface improves rollout efficiency}}: short deterministic interaction chains---\eg, focusing a search box, entering a query, and pressing Enter---can be completed within a single model step, eliminating unnecessary model-environment round trips that would otherwise dominate live-web rollouts. Please refer to Table~\ref{tab:browser-tools} for more details.

\textbf{Context Management.}
Long-horizon web interaction creates a fundamental context-management challenge for multimodal RL agents~\cite{huang2026rethinking}. Each rollout step may contain a full-page screenshot, textual metadata, tool calls, environment feedback, and model reasoning traces. Retaining all screenshots quickly becomes impractical: a 30-step trajectory can exceed the context budget even for 64k-token models. However, preserving every historical screenshot is often unnecessary. Human users do not repeatedly inspect every previous browser state; instead, they rely primarily on recent visual observations together with memory of prior actions, outcomes, and task progress. We adopt the same principle by separating visual grounding from long-term memory. Recent screenshots are retained explicitly for perception, while older interaction history is compressed into textual state information, including environment feedback and the agent's own reasoning traces.

Specifically, at turn $t$, the policy receives the system instruction $s$, the task query $q$, the full sequence of previous model responses $\{y_j\}_{j<t}$, the full sequence of environment feedback strings, and the current browser observation. Each observation is represented as $o_t = (x_t, I_t)$, where $x_t$ contains lightweight textual metadata such as the active tab, URL, and previous-step feedback, while $I_t$ denotes the rendered screenshot.
To bound multimodal context cost, we retain only the most recent $K$ screenshots:
$\mathcal{I}_t =
(I_{\max(0,t-K+1)}, \ldots, I_t)
$. Empirically, retaining only the current screenshot ($K=1$) already achieves strong performance while substantially reducing training cost (Section~\ref{sec:main_results}).
Let $y_j$ denote the model response at turn $j$ which includes both reasoning content and tool calls,
the resulting policy context is: $h_t =
(s, q, o_0, y_0, o_1, y_1, \ldots, o_{t-1}, y_{t-1}, o_t, \mathcal{I}_t)$.

A key design choice is that \textbf{\emph{environment feedback is always preserved}}, even after its corresponding screenshot has been discarded. These feedback strings provide the only explicit execution signal for browser actions, indicating whether interactions succeeded, failed, triggered navigation, opened new tabs, or produced unexpected behavior that may not be visually obvious from the screenshot alone. This becomes particularly important under our multi-tool-call interface, where a single model step may execute several browser actions sequentially and the subsequent turn must identify which sub-actions succeeded and which require recovery or retry.

Unlike prior GUI-agent approaches that compress history into short action summaries~\citep{xu2024aguvis,yang2026guilibra} or executable action code~\citep{gupta2026molmoweb}, we \textbf{\emph{retain the agent's full historical reasoning traces as part of the context}}. We find that these reasoning traces naturally serve as compact textual memory, capturing prior visual observations, action intent, intermediate conclusions, and task progress without requiring all past screenshots to remain in context.

\subsection{Data Preparation and Supervised Fine-tuning}\label{sec:data_sft}

We construct the training corpus from WebGym~\citep{bai2026webgym} through a task-filtering, teacher-rollout, and trajectory-curation pipeline. Starting from 292K raw task instances, we remove tasks that overlap with evaluation benchmarks, subtasks decomposed from parent intents, tasks from long-tail or unstable websites, and near-duplicate intents. To identify near duplicates, we embed task intents with \texttt{Qwen3-Embedding-8B} and apply greedy similarity-based deduplication with a predefined threshold. We use a threshold of $0.99$ to construct the SFT candidate pool, resulting in 15,601 filtered seed tasks. For the RL task pool, we apply the same filtering pipeline with a stricter threshold of $0.95$ to further reduce semantic redundancy, retaining approximately 2.2K tasks spanning diverse real-world websites for RL.

To obtain demonstrations for supervised warm-starting, we rollout a strong open-source teacher model, \texttt{Qwen3-VL-235B-A22B-Thinking}~\citep{bai2025qwen3}, using the agent harness described in Section~\ref{sec:harness}. For each filtered seed task, we sample four independent teacher trajectories and use GPT-4.1 to judge task success based on the final answer, interaction history, and screenshot trajectory. This yields a pool of successful demonstrations with multiple attempts per task, providing both high-quality trajectories and useful signals about task difficulty and rollout diversity.
For SFT, we intentionally curate a small, high-quality subset instead of imitating all successful teacher trajectories. The goal is to provide the small model with sufficient web-interaction competence for productive exploration, while avoiding excessive imitation that may limit the effectiveness of subsequent online RL. By default, we select successful trajectories from the PAE-WebVoyager subset, which densely covers popular real-world websites. For each task group, we retain the shortest successful trajectory. When multiple trajectories have the same length, we use the shorter total response length as a tie-breaker. We also cap the number of tasks per website to improve domain diversity.
This produces our default SFT set of 412 trajectories spanning 70 websites. The same curation rule can be applied to the InSTA-v3 subset to construct a larger SFT set, but in practice we find that the 412-trajectory set is already sufficient for supervised warm-starting small models.

We train small models, such as \texttt{Qwen3-VL-4B-Thinking}, with multi-turn behavioral cloning on the curated trajectories. Each assistant turn is treated as a training target conditioned on the serialized interaction history, including screenshots, reasoning traces, tool calls, and environment feedback. Following standard multi-turn agent-training practice, we apply the loss only to the target assistant response and mask historical context and environment observations.

\subsection{Reward Design and Judge Model}

We design the reward to capture both format correctness and task success, with
details provided in Appendix~\ref{app:reward_design}. The
reward consists of two components.
\textbf{\emph{(1) Format reward.}} First, a format reward checks whether the response follows the required browser-agent protocol: $
r_{\mathrm{fmt}}(y_t) \in \{0,1\}$.
A response receives format credit only if it contains the required thinking
termination tag and a successfully parsed tool call. \textbf{\emph{(2) Task-success reward.}}
For completed trajectories, we use a VLM-as-a-judge reward to evaluate whether
the task has been successfully completed. The judge takes as input the task
instruction, the final response, recent screenshots, and the trajectory history,
including tool calls and environment feedback. Examples of judge inputs and
outputs are provided in Appendix~\ref{app:judge_input_output}. The final reward is assigned at the trajectory level. A trajectory receives a
positive reward only when it both follows the required response format and is
judged successful. It receives a negative reward if it terminates due to repeated
format errors, and receives zero reward otherwise. This design encourages the
agent to produce valid executable actions while optimizing for actual task
completion.

We use GPT-4.1 as the default judge during training. However, relying on a
proprietary judge can be costly and may limit accessibility for the research
community. In our experiments, a typical training run requires 43.2K judge API
calls, costing approximately \$545.5. To reduce this cost and make the pipeline
more accessible, we distill an 8B judge model from 12.5K diverse online rollouts
and GPT-4.1 judge labels. The distilled judge is trained to predict both the
judging rationale and the final success evaluation. We evaluate the effectiveness
of learned judge models in Section~\ref{sec:exp_judge}.

\subsection{Multimodal Multi-turn GRPO}
Since rewards are assigned
only at the trajectory level, we extend GRPO to optimize all assistant
responses across a rollout, assigning the same group-relative advantage to every
turn in the trajectory. For a trajectory $\tau_i$ with $T_i$ assistant turns, we
construct $\{(h_{i,t}, y_{i,t})\}_{t=0}^{T_i-1}$, where $h_{i,t}$ is the managed
multimodal context and $y_{i,t}$ is the assistant response at turn $t$. We use
the same context-management procedure for rollout and optimization, and apply
the loss only to assistant response tokens. 

For each task $q$, we sample a group of $G$ trajectories from the current policy: $\{\tau_i\}_{i=1}^G$, $\tau_i \sim \pi_{\theta_{\mathrm{old}}}$.
Let $R_i = R(\tau_i)$ be the trajectory-level reward.
Following group-relative policy optimization, we compute a normalized group-relative advantage: $A_i = \frac{R_i - \mu_G}{\sigma_G + \epsilon}$,
where $\mu_G = \frac{1}{G}\sum_{j=1}^{G} R_j$ and $\sigma_G$ is the reward standard deviation within the group.
The resulting advantage $A_i$ is assigned to all response tokens from all turns of trajectory $\tau_i$.
We optimize the clipped multi-turn GRPO objective:
{
\small
\[
\mathcal{L}_{\mathrm{MM\text{-}GRPO}}(\theta)
=
-
\frac{1}{G}
\sum_{i=1}^{G}
\sum_{t=0}^{T_i-1}
\frac{
\sum_{k} m_{i,t,k}
\min\left(
\rho_{i,t,k}(\theta) A_i,
\operatorname{clip}
\left(
\rho_{i,t,k}(\theta),
1-\epsilon_{\mathrm{low}},
1+\epsilon_{\mathrm{high}}
\right) A_i
\right)
}{
\max(\sum_k m_{i,t,k}, 1)
},
\]
}
where $\rho_{i,t,k}(\theta) 
=
\frac{\pi_\theta(y_{i,t,k}\mid h_{i,t}, y_{i,t,<k})}{
\pi_{\theta_{\mathrm{old}}}(y_{i,t,k}\mid h_{i,t}, y_{i,t,<k})}$
is the importance ratio for token $k$ in turn $t$ of trajectory $i$, and $m_{i,t,k}$ masks out non-assistant tokens. Note that we do not include the $1/T_i$ normalization factor at the trajectory level, as doing so would downweight longer trajectories and weaken the learning signal for harder tasks that require more interaction steps.

We use asymmetric clipping with $\epsilon_{\mathrm{low}}=0.2$ and 
$\epsilon_{\mathrm{high}}=0.28$, and adopt trajectory-level dynamic sampling 
from~\citep{yu2025dapo}, discarding task groups whose trajectories all receive 
identical rewards (\eg, all-zero or all-one). This filtering removes groups 
dominated by environment failures or trivial tasks, while creating a natural 
curriculum that focuses optimization on tasks that are solvable but not yet 
reliably solved by the current policy. We omit KL and entropy regularization 
to keep the training signal focused on trajectory-level reward.

\section{Experiments}\label{sec:experiments}
\textbf{Training Settings.} We use \texttt{Qwen3-VL-4B-Thinking}~\citep{bai2025qwen3} as the main base model to keep online RL training computationally feasible. 
We first warm-start the model with SFT via LlamaFactory\citep{zheng2024llamafactory} 
for 3 epochs to obtain \texttt{\syssft}. 
Starting from this SFT checkpoint, we further apply our MM-GRPO algorithm in online web environments, using a curated 2.2K training set described in Section~\ref{sec:data_sft}. 
The main RL training runs for 90 iterations, where each iteration consists of online rollouts followed by MM-GRPO updates, requiring about 300 B200 GPU hours in total. 
Across training, we collect approximately 54K online trajectories. 
Since 30-step rollouts are substantially slower, we train \texttt{\sysrl} in two stages: 90 iterations with a maximum of 15 rollout steps, followed by 50 iterations with a maximum of 30 rollout steps. 
We also report the results of a 4B variant trained with our distilled 8B judge model and an 8B variant (\texttt{OpenWebRL-8B}) trained with the same pipeline. 
More training details are provided in Appendix~\ref{app:train_details}.

\textbf{Benchmarks.}
We evaluate on three challenging live web benchmarks: \textit{WebVoyager}~\citep{he2024webvoyager}, 
covering open-domain navigation across popular websites (using the FARA-curated 595-task version~\citep{fara7b2025}); \textit{Online-Mind2Web}~\citep{xue2025an}, comprising 300 long-horizon tasks across 136 websites; and \textit{DeepShop}~\citep{lyu2025deepshop}, 
targeting realistic shopping tasks with multi-constraint product selection. We follow 
each benchmark's standard evaluation protocol. Full details are in 
Appendix~\ref{app:benchmark_details}.

\textbf{Baselines.} We compare our models with both proprietary and open-source baselines.
The proprietary baselines include GPT-5, Gemini-3-Flash, o3, OpenAI CUA, and Gemini CUA.
The open-source baselines include Holo1-7B~\citep{andreux2025surfer}, UI-TARS-1.5-7B~\citep{qin2025ui}, GLM-4.1V-9B-Thinking~\citep{hong2025glm}, FARA-7B~\citep{fara7b2025}, MolmoWeb-4B/8B~\citep{gupta2026molmoweb}, Qwen3-VL-4B-Thinking, and Qwen3-VL-235B-A22B-Thinking~\citep{bai2025qwen3}.

\textbf{Evaluation Metrics.} Our primary metric is the \emph{official success rate}, following the standard evaluation protocol used in prior work~\cite{fara7b2025,gupta2026molmoweb}. This protocol relies on Browser-Use Stealth Browsers\footnote{\url{https://browser-use.com/stealth-browsers}}, a managed cloud-browser service that provides CAPTCHA solving and stable browser sessions, reducing failures caused by website blocking and session instability during live-web evaluation.
However, Browser-Use Service introduces a paid third-party dependency, which increases evaluation cost and makes exact reproduction less accessible for the academic community. To improve transparency, we also report ``\textit{success rate w/o aborted tasks}''---the success rate excluding tasks with non-agent failures such as blocked pages or disconnected browser sessions, providing an estimate of agent performance without using Browser-Use Service.

\begin{table*}[t]
\centering
\caption{Official success rates (\%) across three open-web benchmarks. * marks numbers reported in FARA~\citep{fara7b2025}; $^{\dagger}$ marks numbers reported in MolmoWeb~\citep{gupta2026molmoweb}.}
\label{tab:browser-results}
\small
\setlength{\tabcolsep}{2pt}
\resizebox{1\textwidth}{!}{
\begin{tabular}{lccccc>{\columncolor{blue!10}}c}
\toprule
\textbf{Model Name} & \textbf{\# Steps} & \textbf{\# Tasks} & \textbf{WebVoyager} & \textbf{Online-Mind2Web} & \textbf{DeepShop} & \textbf{Average} \\
\midrule
\multicolumn{7}{l}{\textit{Proprietary Models}} \\
\midrule
GPT-5 (Axtree)$^{\dagger}$              & 30   & -- & 70.6 & 41.9 & 40.7 & 51.1 \\
Gemini-3-flash (Axtree)$^{\dagger}$     & 30   & -- & 74.4 & 34.8 & 45.1 & 51.4 \\
Gemini-3-flash (Axtree)$^{\dagger}$     & 100  & -- & 85.6 & 44.8 & 55.3 & 61.9 \\
GPT-4o (SoM)*                            & 100  & -- & 65.1 & 34.6 & 16.0 & 38.6 \\
o3 (SoM)*                                & 100  & -- & 79.3 & 55.4 & 49.7 & 61.5 \\
GPT-5 (SoM)*                             & 100  & -- & \textbf{90.6} & 57.7 & 49.1 & 65.8 \\
OpenAI computer-use-preview*             & 100  & -- & 70.9 & 58.3 & 24.7 & 51.3 \\
Gemini computer-use-preview$^{\dagger}$  & 100  & -- & \underline{88.6} & 57.3 & 62.0 & \textbf{69.3} \\
\midrule
\multicolumn{7}{l}{\textit{Open-Source Models}} \\
\midrule
Holo1-7B$^{\dagger}$                     & 30   & $>$15.6k  & 55.4 & --   & --   & --   \\
UI-TARS-1.5-7B*                          & 100  & --      & 66.4 & 31.3 & 11.6 & 36.4 \\
GLM-4.1V-9B-Thinking*                    & 100  & --      & 66.8 & 33.9 & 32.0 & 44.2 \\
Fara-7B*                                 & 100  & $>$123.2k & 73.5 & 34.1 & 26.2 & 44.6 \\
MolmoWeb-4B$^{\dagger}$                  & 100  & $>$278.5k & 75.2 & 31.3 & 35.6 & 47.4 \\
MolmoWeb-8B$^{\dagger}$                  & 100  & $>$278.5k & 78.2 & 35.3 & 42.3 & 51.9 \\
Qwen3-VL-4B-Thinking                     & 30   & --      & 52.6 & 32.0 & 33.3 & 39.3 \\
Qwen3-VL-8B-Thinking                     & 30   & --      & 61.3 & 38.7 & 44.0 &  48.0  \\
Qwen3-VL-235B-A22B-Thinking              & 30   & --      & 66.4 & 63.7 & 56.7 & 62.3 \\

\midrule
\multicolumn{7}{@{}l}{\textit{Ours: 4B backbone}} \\
\midrule
\rowcolor{gray!8}
\textbf{OpenWebRL-4B-SFT}                  & 30 & 0.4k & 60.2 & 47.0 & 48.7 & \avgcell{52.0} \\
\rowcolor{gray!8}
\textbf{OpenWebRL-4B}                      & 30 & 2.2k & 74.1 & 67.0 & 64.0 & \avgcell{68.4} \\
\rowcolor{gray!8}
\makecell[l]{\textbf{OpenWebRL-4B} \textbf{w/ OpenWebRL-Judge-8B}}
                                             & 30 & 2.2k & 68.9 & \underline{67.3} & \textbf{68.7} & \avgcell{68.3} \\

\midrule
\multicolumn{7}{@{}l}{\textit{Ours: 8B backbone}} \\
\midrule
\rowcolor{gray!8}
\textbf{OpenWebRL-8B-SFT}                  & 30 & 0.9k & 66.2 & 54.0 & 50.0 & \avgcell{56.7} \\
\rowcolor{gray!8}
\textbf{OpenWebRL-8B}                      & 30 & 2.2k & 73.8 & 67.0 & \underline{65.3} & \avgcell{68.7} \\

\rowcolor{gray!8}
\textbf{OpenWebRL-8B}                      & 50 & 2.2k & 74.6 & \textbf{69.7} & 63.3 & \avgcell{\underline{69.2}} \\

\bottomrule
\end{tabular}}
\end{table*}

\subsection{Main Results}
\label{sec:main_results}

\textbf{OpenWebRL establishes a new open-source state of the art and is competitive with proprietary web agents.}
Table~\ref{tab:browser-results} reports official success rates on three live-web benchmarks.
With only a 4B backbone, a 30-step evaluation budget, and a much smaller training set than prior open-source systems, {\sysrl} achieves an average success rate of $68.4\%$, substantially outperforming existing open-source agents.
The gains of {\sysrl} are particularly pronounced on long-horizon benchmarks, \ie, Online-Mind2Web and DeepShop, as indicated by Appendix~\ref{app:task_difficulty}. {\sysrl} outperforms FARA-7B by $+32.9$ and $+37.8$ points, and surpasses MolmoWeb-8B by $+31.7$ and $+21.7$ points on the two benchmarks, respectively.
Scaling the backbone to 8B further improves the average success rate to $68.7\%$ under the same 30-step evaluation budget. \textit{Increasing the evaluation budget to 50 steps yields a slightly higher average success rate of $69.2\%$ for OpenWebRL-8B, suggesting that the model can generalize to longer interaction horizons} despite being trained with 30-step SFT and RL budgets. However, the 50-step setting also incurs substantially higher computation and wall-clock cost, so we use 30 steps as the default evaluation setting.
Notably, both OpenWebRL-4B and OpenWebRL-8B are highly competitive with proprietary systems.
On Online-Mind2Web and DeepShop, they surpass several closed-source agents, including GPT-5, Gemini-3-Flash, GPT-4o, o3, and the OpenAI/Gemini computer-use agents.
These results show that effective online RL can substantially improve open-source VLMs for live-web interaction, enabling compact 4B--8B models to compete with much larger proprietary systems on challenging live-web tasks.

\textbf{Both SFT and MM-GRPO contribute to the final gains.}
Table~\ref{tab:browser-results} shows that supervised fine-tuning provides a strong warm start for both model scales, while MM-GRPO contributes even larger improvements in the subsequent online RL stage.
For the 4B backbone, SFT improves the average success rate from $39.3\%$ to $52.0\%$, while MM-GRPO further increases it to $68.4\%$, yielding gains of $+16.4$ points over SFT and $+29.1$ points over the base model.
We observe a similar trend for the 8B backbone: SFT improves the average success rate from $48.0\%$ to $56.7\%$, and MM-GRPO further raises it to $68.7\%$, giving gains of $+12.0$ points over SFT and $+20.7$ points over the original backbone.
The improvements are consistent across all three benchmarks, suggesting that SFT provides the interaction competence needed for exploration, while online RL is the key driver of the final performance gains.

\subsection{Learning Dynamics of MM-GRPO}
\begin{figure}[t]
    \centering
    \includegraphics[width=1\linewidth]{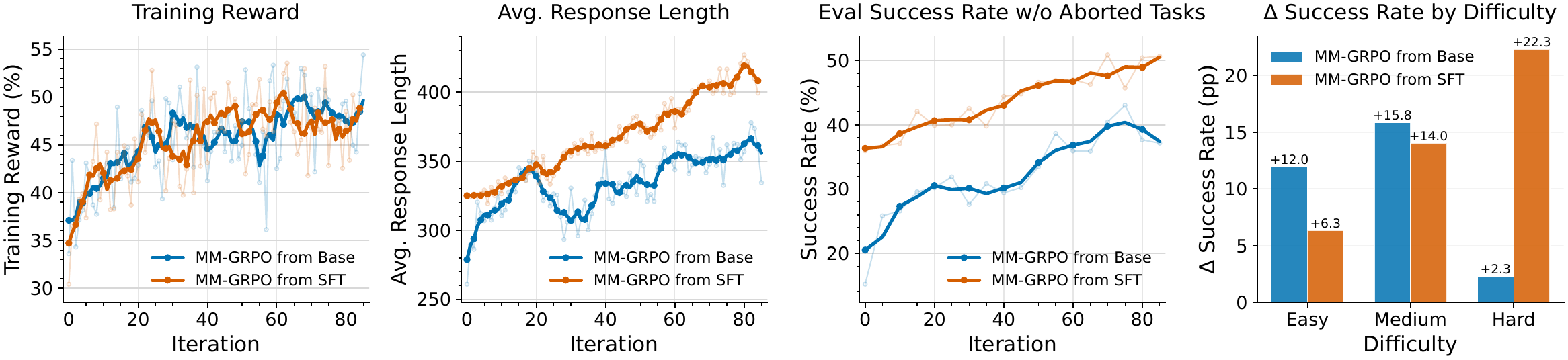}
    \vspace{-12pt}
    \caption{Comparison of MM-GRPO training from SFT and base-model initializations. (a) Average training reward and (b) average response length measured on training rollouts. (c) Evaluation success rate (excluding aborted tasks) and (d) success rate improvement across difficulty splits measured on Online-Mind2Web.}
    \label{fig:learning_curve_sft_base}
\end{figure}

\textbf{SFT initialization leads to faster and more effective RL.}
Figure~\ref{fig:learning_curve_sft_base} compares MM-GRPO training initialized from the base model (\texttt{Qwen3-VL-4B-Thinking}) and from the SFT checkpoint on Online-Mind2Web benchmark.
Although both runs eventually reach similar training rewards and follow a similar evaluation trend, the SFT-initialized run maintains a clear $\sim$10\% advantage in evaluation success rate throughout training.
This gap suggests that supervised warm-starting does more than simply accelerate optimization: it places the policy in a better region of the behavior space, enabling more effective online exploration.
Overall, MM-GRPO produces stable improvements from both initializations, and the SFT-initialized policy continues to improve even after 80 iterations.

\textbf{The benefit of SFT initialization is especially pronounced on harder tasks.}
Figure~\ref{fig:learning_curve_sft_base}(d) further breaks down performance by task difficulty.
MM-GRPO from the SFT checkpoint improves success across all difficulty splits, with the largest gain on hard tasks (+22.3 points).
In contrast, RL from the base model improves only marginally on hard tasks (+2.3 points), with most of its gains concentrated on the easy and medium splits.
Together, these results validate the effectiveness and necessity of warm-starting online RL from a SFT model rather than training directly from the base model.

\begin{figure}
    \centering
    \includegraphics[width=1\linewidth,trim=5 0 5 25,clip]{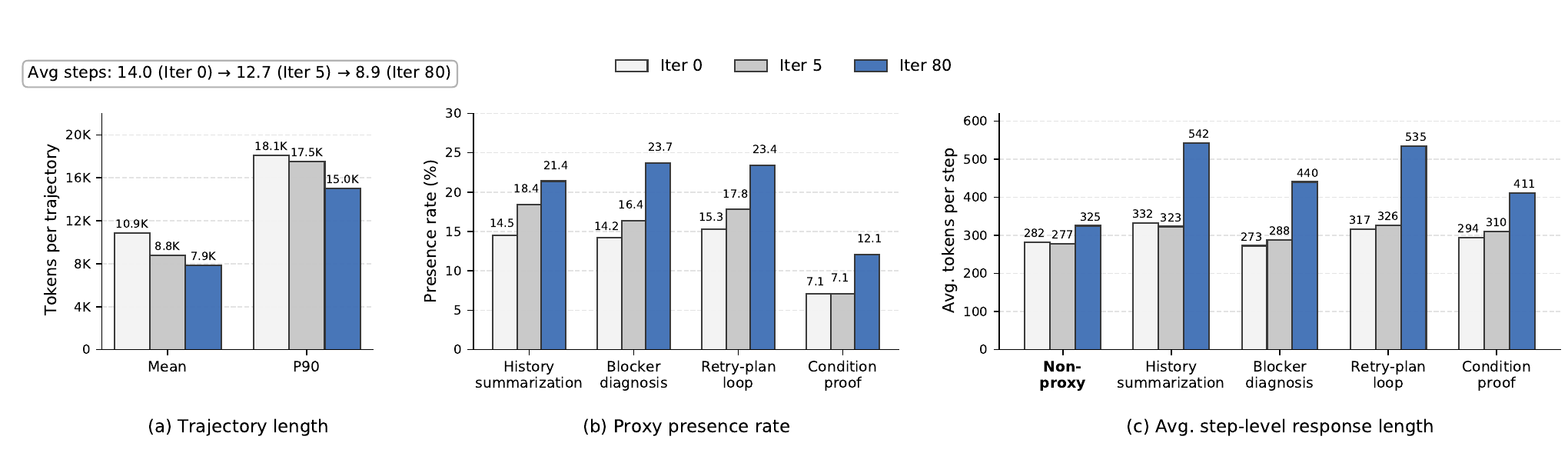}
    \vspace{-20pt}
    \caption{Diagnosing response-length growth during MM-GRPO.
(a) Trajectory length, measured by the mean and P90 (90th percentile), together with the average number of interaction steps.
(b) Step-level proxy presence rates.
(c) Average step-level response length conditioned on proxy type, including no-proxy steps.
 }
    \label{fig:reasoning_pattern_change}
\end{figure}

\textbf{MM-GRPO drives targeted rather than uniform reasoning expansion.} As shown in Figure \ref{fig:learning_curve_sft_base} (b), the model produces longer responses during MM-GRPO training. We further analyze how response length changes in Figure~\ref{fig:reasoning_pattern_change}. Interestingly, \textbf{the overall trajectory length does not increase.} Instead, the average number of interaction steps decreases from 14.0 at iteration 0 to 8.9 at iteration 80, while the average and 90th-percentile trajectory lengths drop from 10.9K and 18.1K tokens to 7.9K and 15.0K tokens, respectively.
To better understand the source of the increased response length, we apply lexical proxy detection to model outputs on Online-Mind2Web to identify several step-level reasoning patterns (see Appendix~\ref{app:proxy_analysis} for details). As shown in Figure~\ref{fig:reasoning_pattern_change}(b) and (c), MM-GRPO increases both the frequency and the conditional length of several recurring reasoning patterns, including history summarization, blocker diagnosis, retry-plan reasoning, and condition-proof reasoning. For instance, the step-level presence rate rises from 14.5\% to 21.4\% for history summarization, from 14.2\% to 23.7\% for blocker diagnosis. The average response length for proxy-bearing steps also grows substantially, from 332 to 542 tokens for history summarization, from 273 to 440 tokens for blocker diagnosis. By comparison, non-proxy steps remain much more stable, increasing only from 282 to 325 tokens on average. \textbf{\emph{These results suggest that response-length growth during RL is not a uniform expansion across all responses; instead, the model allocates additional verbosity and reasoning patterns selectively to important steps.} }

\subsection{Test-time Scaling} \label{sec:test-scaling}
\begin{figure}
    \centering
    \includegraphics[width=0.31\linewidth]{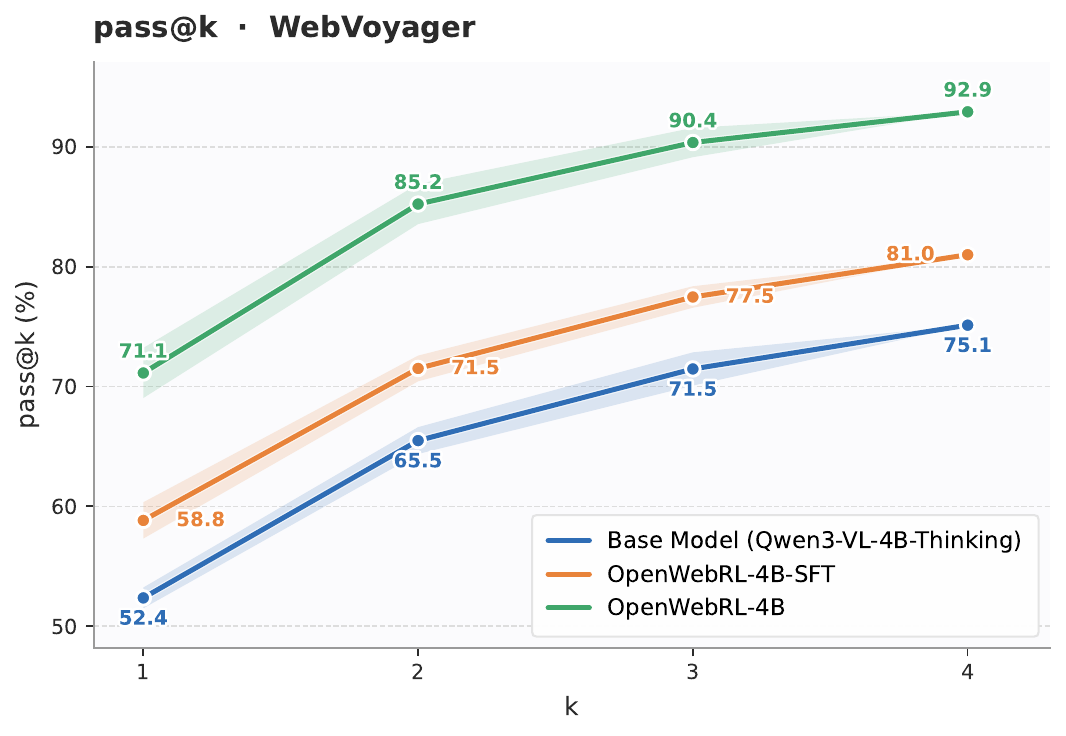}
    \includegraphics[width=0.31\linewidth]{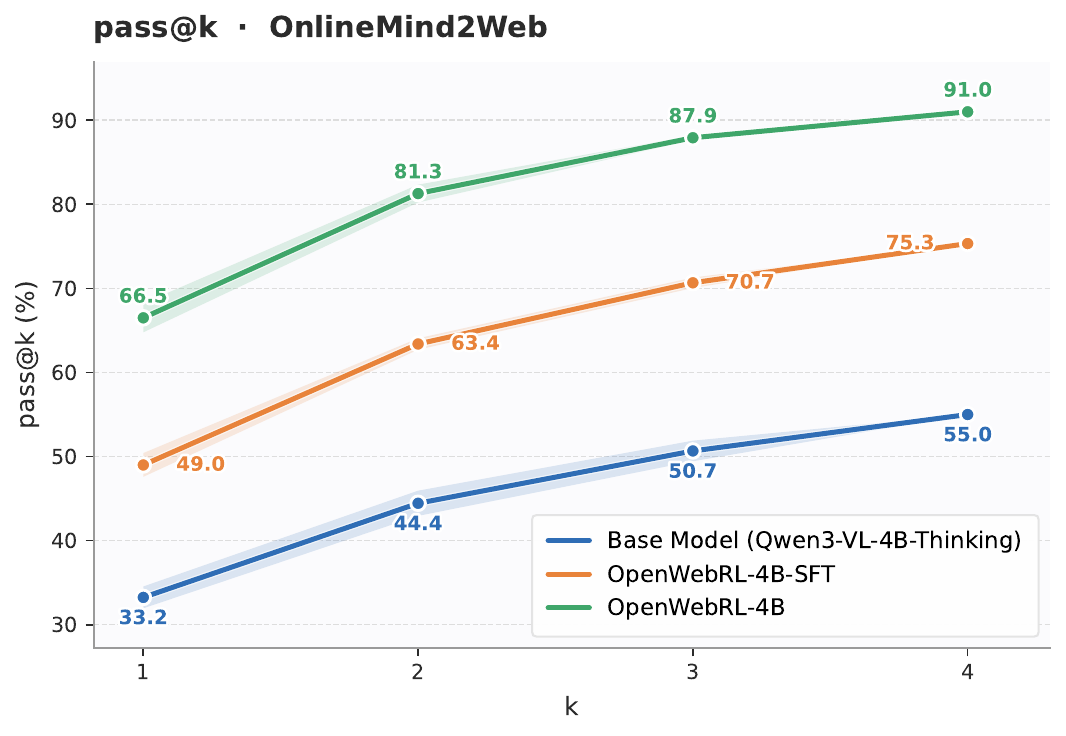}
    \includegraphics[width=0.31\linewidth]{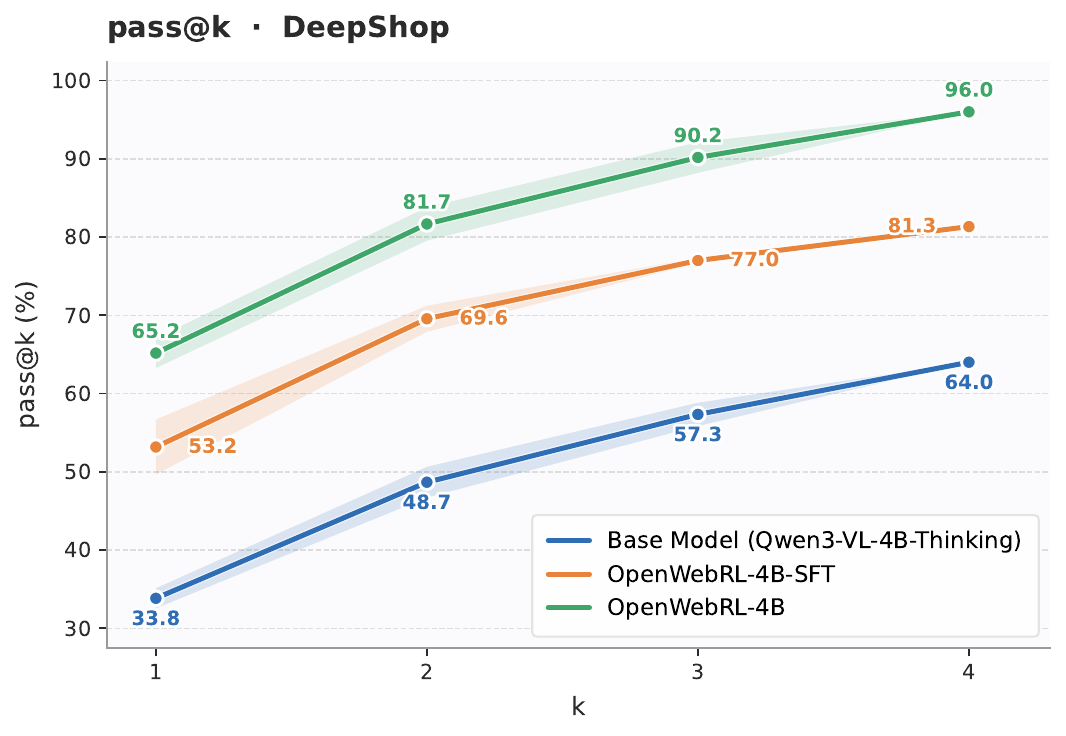}
    \caption{Pass@k performance of OpenWebRL-4B and baselines on three online benchmarks. }
    \label{fig:test_time_scaling}
\end{figure}

We further study whether visual web agents can benefit from additional test-time computation.
Specifically, we evaluate \textit{pass@k}, where each task is attempted with $k$ independent rollouts and is counted as successful if at least one rollout succeeds.
We use $k \in \{1,2,3,4\}$ and keep the inference setup fixed across models, with a maximum rollout length of $30$ steps.
Figure~\ref{fig:test_time_scaling} compares the base model (Qwen3-VL-4B-Thinking), OpenWebRL-4B-SFT, and OpenWebRL-4B on WebVoyager, Online-Mind2Web, and DeepShop.
The results show that live-web agent performance can be scaled by sampling multiple independent attempts.
However, \textbf{OpenWebRL-4B exhibits a consistently stronger pass@k curve than both the base and SFT models across all three benchmarks.}
This indicates that online RL improves not only single-rollout success, but also the probability that repeated attempts discover a successful trajectory.
Notably, OpenWebRL-4B achieves over $90\%$ pass@4 official success rate across the three live-web benchmarks, substantially outperforming MolmoWeb-8B~\citep{gupta2026molmoweb}, which reports around $60\%$ pass@4 success rate on Online-Mind2Web.
The strong pass@k scaling suggests that OpenWebRL-4B learns not only more accurate individual actions, but also a richer distribution of viable interaction strategies that can be exploited through multiple test-time attempts.

\subsection{Evaluation of OpenWebRL-Judge}
\label{sec:exp_judge}

\begin{figure}[t]
    \centering
    \begin{minipage}[t]{0.54\textwidth}
        \centering
        \captionof{figure}{Training and evaluation curve for RL with different judges.}
        \includegraphics[width=\linewidth]{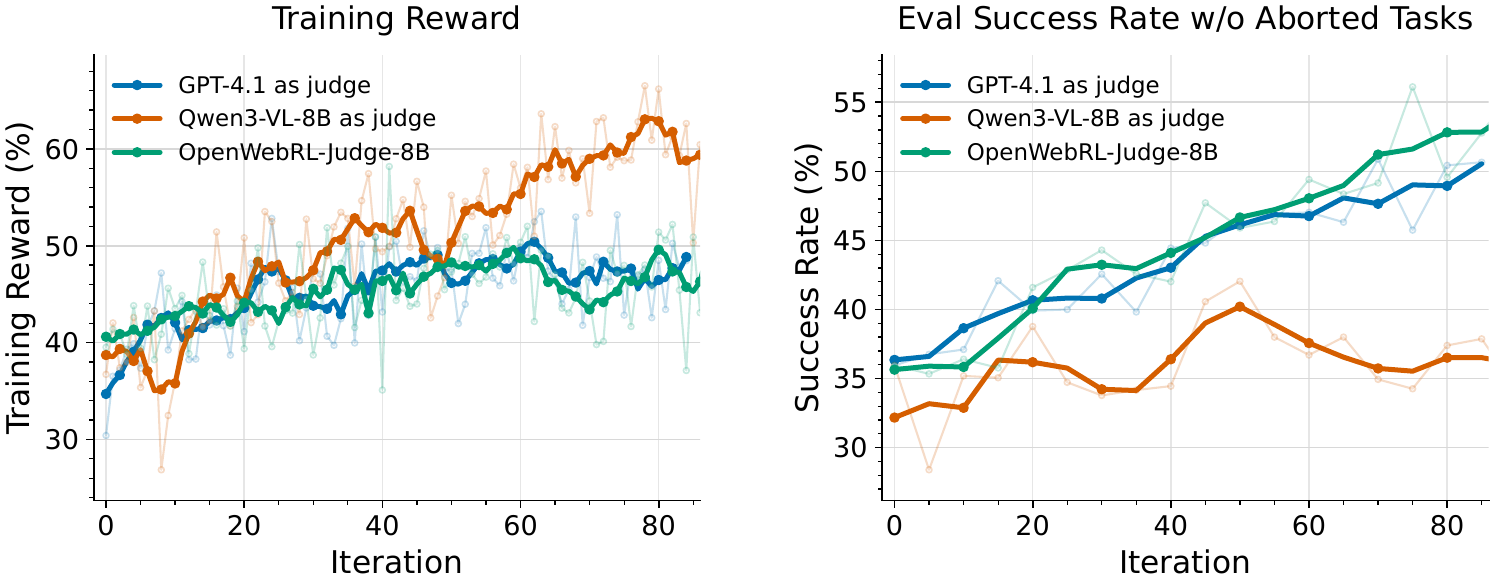}
        \label{fig:rl_with_judge}
    \end{minipage}
    \hfill
    \begin{minipage}[t]{0.45\textwidth}
        \centering
        \captionof{table}{Judge model evaluation results. Oracle is labeled by GPT-4.1. All numbers are percentages.}
        \label{tab:judge_results}
        \resizebox{\linewidth}{!}{%
        \begin{tabular}{lcccc}
        \toprule
        \textbf{Model} & \textbf{Accuracy} & \textbf{Precision} & \textbf{Recall} & \textbf{F1} \\
        \midrule
        o4-mini & 76.0 & \textbf{95.2} & 61.7 & 74.9 \\
        gpt-4o & 85.6 & 83.6 & 93.4 & \underline{88.3} \\
        WebJudge-7B~\cite{xue2025an} & 71.6 & 67.9 & \textbf{96.9} & 79.8  \\
        Qwen3-VL-4B-Instruct   & 76.8 & 75.7 & 89.0 & 81.8 \\
        Qwen3-VL-8B-Instruct   & 80.6 & 78.6 & 91.7 & 84.7 \\
        Qwen3-VL-32B-Instruct   & \underline{85.8} & 87.4 & 88.3 & 87.8 \\
        \midrule
        OpenWebRL-Judge-4B     & 85.2 & 86.1 & 89.7 & 87.8 \\
        OpenWebRL-Judge-8B     & \textbf{89.8} & \underline{89.5} & \underline{94.8} & \textbf{92.1} \\
        \bottomrule
        \end{tabular}   }
    \end{minipage}
\end{figure}

\textbf{The distilled OpenWebRL-Judge-8B provides effective training signals for stable online RL.} As shown in Table~\ref{tab:browser-results}, \textit{{\sysrl} w/ OpenWebRL-Judge-8B} achieves strong performance among open-source agents on both Online-Mind2Web (67.3) and DeepShop (68.7). Its overall average score (68.3) is nearly identical to that of the GPT-4.1-Judge variant (68.4), despite relying entirely on a distilled open-source reward model during RL training, substantially reducing both the cost and dependency associated with proprietary API calls.
We further compare RL training dynamics in Figure~\ref{fig:rl_with_judge}. Both the training reward and the evaluation success rate w/o aborted tasks on Online-Mind2Web show stable optimization trends that closely match those obtained with GPT-4.1 supervision. In contrast, using \emph{Qwen3-VL-8B as the judge model leads to clear reward hacking behavior}~\cite{gao2023scaling,yang2024regularizing,miao2024inform}, producing higher training rewards but substantially lower evaluation success rates. These results further demonstrate that OpenWebRL-Judge-8B provides reliable and effective reward signals for stable online RL training.

\textbf{OpenWebRL-Judge-8B closely matches GPT-4.1 on held-out trajectory evaluation.} To evaluate the judge model independently from RL training, we construct a held-out set of 500 trajectory rollouts collected from different training stages and annotate them with GPT-4.1. We then compare several judge models by measuring their agreement with these GPT-4.1 annotations. As shown in Table~\ref{tab:judge_results}, OpenWebRL-Judge-8B achieves the strongest alignment with GPT-4.1, reaching 89.8\% accuracy and a 92.1\% F1 score. It outperforms strong baselines, including WebJudge-7B~\cite{xue2025an}, Qwen3-VL-32B, and GPT-4o. While some baselines achieve strong precision or recall individually, their overall accuracy and F1 scores remain lower, suggesting that our distilled judge captures web-task success signals more reliably than both general-purpose VLM judges and prior web-specific judge models in this setting.

\begin{table*}[t]
\centering
\caption{Ablation study on rollout length and context management strategies for online RL.}
\label{tab:gui:ablation_study}
\small
\setlength{\tabcolsep}{4pt}
\resizebox{0.8\textwidth}{!}{
\begin{tabular}{lccc}
\toprule
\textbf{System} & \textbf{WebVoyager} & \textbf{Online-Mind2Web} & \textbf{DeepShop} \\
\midrule
OpenWebRL-4B & \textbf{74.1} & \textbf{67.0} & \textbf{64.0} \\
\midrule
\multicolumn{4}{l}{\textit{Rollout length ablations}} \\
\quad RL w/ 30-Step Rollout Only
& 66.7~{\scriptsize(\textcolor{red}{$\downarrow$ -7.4})}
& 65.4~{\scriptsize(\textcolor{red}{$\downarrow$ -1.6})}
& 63.3~{\scriptsize(\textcolor{red}{$\downarrow$ -0.7})} \\
\quad RL w/ 15-Step Rollout Only
& 70.1~{\scriptsize(\textcolor{red}{$\downarrow$ -4.0})}
& 65.0~{\scriptsize(\textcolor{red}{$\downarrow$ -2.0})}
& 63.3~{\scriptsize(\textcolor{red}{$\downarrow$ -0.7})} \\
\quad RL w/ 10-Step Rollout Only
& 70.6~{\scriptsize(\textcolor{red}{$\downarrow$ -3.5})}
& 60.7~{\scriptsize(\textcolor{red}{$\downarrow$ -6.3})}
& 57.3~{\scriptsize(\textcolor{red}{$\downarrow$ -6.7})} \\
\midrule
\multicolumn{4}{l}{\textit{Context management ablations with 15-step rollouts}} \\
\quad \quad w/ Recent Two Screenshots
& 68.2~{\scriptsize(\textcolor{red}{$\downarrow$ -1.9})}
& 65.3~{\scriptsize(\textcolor{green}{$\uparrow$ +0.3})}
& 59.3~{\scriptsize(\textcolor{red}{$\downarrow$ -4.0})} \\
\quad \quad w/o Textual Environment Feedback
& 64.9~{\scriptsize(\textcolor{red}{$\downarrow$ -5.2})}
& 57.0~{\scriptsize(\textcolor{red}{$\downarrow$ -8.0})}
& 56.7~{\scriptsize(\textcolor{red}{$\downarrow$ -6.6})} \\
\quad \quad w/o Historical Reasoning
& 55.5~{\scriptsize(\textcolor{red}{$\downarrow$ -14.6})}
& 41.3~{\scriptsize(\textcolor{red}{$\downarrow$ -23.7})}
& 54.7~{\scriptsize(\textcolor{red}{$\downarrow$ -8.6})} \\
\bottomrule
\end{tabular}}
\end{table*}

\subsection{Ablations on Rollout and Context Management}
\label{sec:ablation}

Here we ablate key design choices in {\sysrl}: (i) the rollout-length curriculum used in {\sysrl}, where we first run online RL with a
15-step horizon and then continue with a 30-step horizon; and (ii) the three
context management strategies, including textual environment feedback, historical
reasoning, and the number of recent screenshots retained in context. To ensure a controlled comparison, all variants are initialized from the same SFT checkpoint and use the identical RL training recipe, with only the ablated factor changed. For cost efficiency, the context-management ablations are trained using only 15-step rollouts and are compared against the ``\textit{RL w/ 15-Step Rollout Only}'' baseline. Table~\ref{tab:gui:ablation_study} reports the results.

\textbf{Rollout-length curriculum.}
{\sysrl} outperforms all fixed-budget RL variants. Training with only 30-step rollouts leads to performance drops of $7.4$, $1.6$, and $0.7$ points on WebVoyager, Online-Mind2Web, and DeepShop, respectively. Using only 15-step rollouts performs better than starting directly with 30-step rollouts, but still underperforms the full curriculum by $4.0$, $2.0$, and $0.7$ points. Moreover, using an overly short horizon, such as 10-step rollouts, substantially hurts performance on Online-Mind2Web and DeepShop ($-6.3$ and $-6.7$ points), both of which require longer interaction horizons to complete tasks successfully. These results suggest that a medium-horizon training stage helps stabilize early exploration, while longer-horizon rollouts improve the policy’s ability to handle tasks requiring extended interactions. Overall, combining 15-step and 30-step training stages is more effective than using a fixed rollout budget throughout training.

\textbf{Textual environment feedback provides lightweight but informative signals.}
Removing this feedback reduces RL performance by $5.2$, $8.0$, and $6.6$ points relative to the 15-step rollout baseline.
This implies that the feedback provides explicit information about action execution that is not always inferable from the screenshot alone. Without it, the policy is less reliable in deciding whether to retry, recover from errors, or adjust strategy, particularly in longer-horizon and more failure-prone settings such as Online-Mind2Web.

\textbf{Historical reasoning provides critical contextual signals.}
Removing historical reasoning yields the largest degradation among all ablations, reducing performance by $14.6$, $23.7$, and $8.6$ points relative to the 15-step rollout baseline.
We conjecture that these reasoning traces can serve as a compact memory of the high-level plan, interpretations of past observations, attempted actions, and their outcomes.
Without this information, the model must effectively replan from scratch at each step and loses access to accumulated context, which is particularly detrimental in long-horizon web navigation.

\textbf{Keeping more recent screenshots does not consistently help.}
By default, the model retains only the most recent screenshot in context. Increasing this to the two most recent screenshots does not yield consistent gains, changing performance by $-1.9$, $+0.3$, and $-4.0$ points.
This suggests that historical reasoning already captures most useful information from earlier visual observations, while additional screenshots mainly increase context length and add visual token overhead. Beyond the latest frame, visual history provides limited benefit and may even degrade performance. It also significantly increases training cost, extending runtime from approximately 240 GPU hours to 400 GPU hours.

\subsection{Ablations on MM-GRPO Designs}
\label{sec:ablation_mmgrpo}

\begin{figure}
    \centering
    \includegraphics[width=0.65\linewidth]{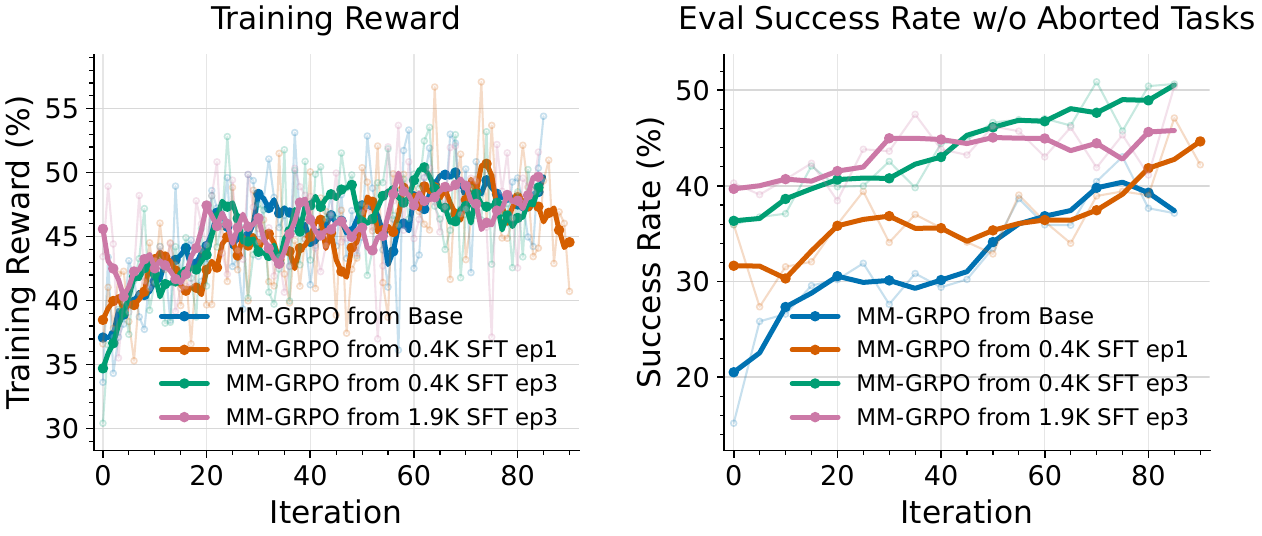}
    \caption{Comparison of MM-GRPO training under different supervised warm-start initializations.
   }
    \label{fig:sft_init}
\end{figure}

\textbf{Online RL benefits most from a balanced SFT warm start.}
We first study how supervised warm-starting affects subsequent online RL. Figure~\ref{fig:sft_init} compares MM-GRPO training from four initial 4B policies: the base model without SFT, a lightweight SFT model trained on 0.4K trajectories for 1 epoch, our default SFT model trained on the same 0.4K trajectories for 3 epochs, and a larger-data SFT model trained on 1.9K trajectories for 3 epochs. The 1.9K trajectories are collected with the same teacher rollout pipeline, but retain a broader and less aggressively curated task distribution from WebGym. The results show that supervised initialization is crucial, i.e., all SFT initializations outperform the base model initialization, but \textit{stronger warm start does not necessarily lead to better online RL}. The 1.9K/3-epoch initialization underperforms the default setting after online RL. Although it starts from a relatively strong policy, its improvement quickly saturates, and its final evaluation performance remains below that of the 0.4K/3-epoch initialization. We hypothesize that heavier imitation training, especially on a less curated task distribution, may reduce policy plasticity or bias the policy toward behaviors that are less amenable to online refinement. In contrast, the default initialization gives the policy enough competence to explore effectively while still allowing RL to further adapt it through online trial and error.

\begin{figure}
    \centering
    \includegraphics[width=1\linewidth]{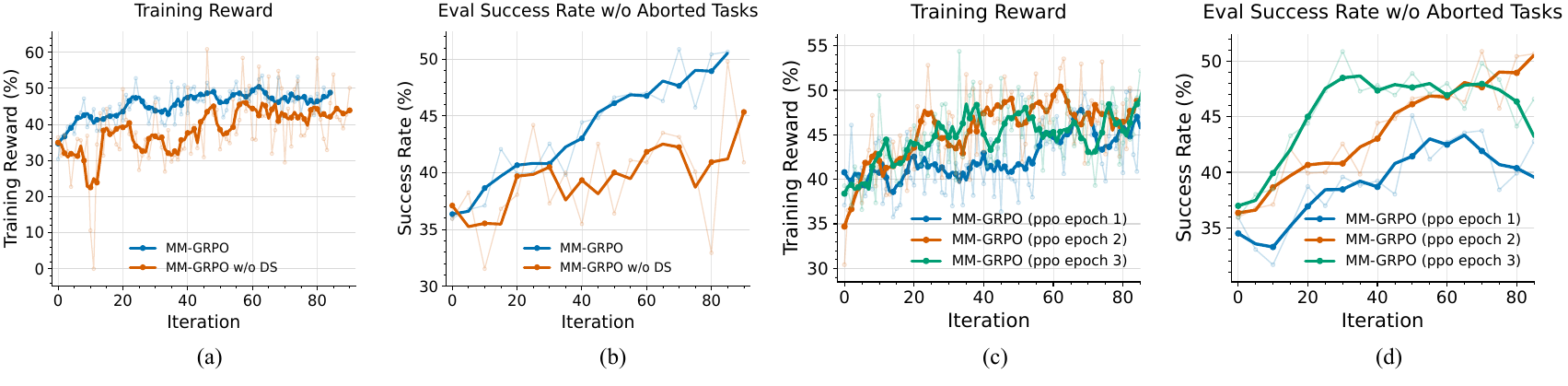}
    \vspace{-20pt}
    \caption{Ablation study on (a)(b) dynamic sampling (DS) and (c)(d) PPO epoch for MM-GRPO.}
\label{fig:ablations_dynamicsampling_ppoepoch}
\end{figure}

\textbf{Trajectory dynamic sampling improves training stability and optimization efficiency.} As shown in Figure~\ref{fig:ablations_dynamicsampling_ppoepoch}(a) and (b), removing trajectory-level dynamic sampling leads to noticeably less stable training dynamics in both training reward and evaluation success rate. With dynamic sampling, training collects a fixed number of effective groups by filtering out zero-reward-variance groups, whereas the variant without dynamic sampling always collects a fixed number of groups regardless of reward variance. As a result, the effective batch size can vary significantly across iterations without dynamic sampling, leading to noisier policy updates and less stable optimization. In contrast, dynamic sampling stabilizes training and enables faster performance improvement in MM-GRPO.

\textbf{Moderate PPO epochs provide a balance between data efficiency and stable optimization.} PPO epochs control how many times the policy is updated on the same rollout batch before collecting new trajectories, creating a trade-off between data efficiency and off-policyness. As shown in Figure~\ref{fig:ablations_dynamicsampling_ppoepoch}(c) and (d), using only one PPO epoch leads to slower optimization and worse evaluation performance, suggesting insufficient updates for each rollout batch. Increasing the PPO epochs to 2 substantially improves both training reward and evaluation success rate, achieving the best overall performance. However, increasing the PPO epochs to 3 results in less stable evaluation behavior: although training rewards remain high, the evaluation success rate peaks early and later declines. This suggests that excessively reusing the same rollout data increases off-policyness and can lead to over-optimization, while a moderate number of PPO epochs better balances sample efficiency and stable online RL training.

\section{Additional Analysis}
\label{sec:analysis}

We provide additional diagnostics in the appendix to provide further insight into the main results.
\begin{itemize}
    \item \textbf{Appendix~\ref{app:task_difficulty} analyzes benchmark difficulty} and shows that Online-Mind2Web and DeepShop require longer trajectories and have lower success rates than WebVoyager, making the gains of OpenWebRL-4B on these benchmarks especially informative.
    
    \item \textbf{Appendix~\ref{app:error-analysis} further analyzes 100 failed trajectories from OpenWebRL-4B.} Access and environment issues account for the largest share of failures (51\%), followed by reasoning and constraint-tracking limitations (27\%), visual grounding and interaction errors (13\%), and task definition or judge issues (9\%). These results suggest that future progress will require not only stronger policies, but also more robust open-web agent infrastructure.
\end{itemize}

\section{Conclusion}
We introduced \textbf{OpenWebRL}, an open framework for training visual web agents with online multi-turn RL on live websites. Built on this framework, \textbf{OpenWebRL-4B} achieves strong performance on challenging live-web benchmarks, demonstrating that effective web-agent training can emerge from relatively modest supervised initialization and scalable online interaction, rather than relying on large-scale demonstration datasets.
Through systematic analysis, we identify several key ingredients that make online RL effective for visual web agents, including supervised warm start, textual environment feedback, historical reasoning for long-horizon memory, rollout-length curricula, and reliable trajectory-level success judging. 
We hope OpenWebRL serves as an actionable and reproducible foundation for future research on open, cost-efficient, and scalable web-agent training, and helps narrow the gap between open and proprietary systems for long-horizon interaction on the live web.

\bibliographystyle{plain}  
\bibliography{main}


\newpage
\appendix

\section{Implementation Details}\label{app:impelmentation_details}

\subsection{Robust Web Environment Infra}
\label{app:env_infra}
To support browser-agent training on live websites, we implement a robust web environment that explicitly handles the instability of the open web. Unlike self-hosted benchmark sites, live websites exhibit frequent layout changes, asynchronous loading, pop-ups, redirects, anti-automation defenses, network errors, and non-deterministic page states.

\paragraph{Sandboxed Execution.} The environment is built on Playwright with a Chromium backend and supports sandboxed execution by default. Each rollout runs inside an isolated Kubernetes sandbox~\cite{peng2026orchard} with its own browser environment server, preventing crashes, stale cookies, memory leaks, and site-specific side effects from contaminating other trajectories.

\paragraph{Timeouts and Recovery.} The browser environment uses separate timeouts for initialization, interaction, and screenshot capture. Initial navigation is retried multiple times to handle common live-site failures such as HTTP/2 errors, connection resets, slow scripts, and transient bot-detection behavior.

\paragraph{Failure Attribution and Diagnostics.}
The rollout layer separates model failures from environment and infrastructure failures using explicit termination reasons, including task completion, maximum-step exhaustion, generation length limits, formatting errors, environment step errors, sandbox failures, and initialization failures. The system further records diagnostics such as server latency, memory usage, uptime, request counts, and tracebacks, enabling post-hoc analysis of whether a low reward reflects model behavior or external factors such as blocking, network instability, or server failure. Network-related initialization errors can also update a host blacklist, allowing future training to avoid websites that are repeatedly unreachable or automation-hostile.

\begin{table}[t]
\centering
\small
\caption{Examples of action-level environment feedback returned by the browser environment. Feedback exposes both execution status and observable browser-state changes.}
\begin{tabular}{p{0.12\linewidth}p{0.36\linewidth}p{0.38\linewidth}}
\toprule
Action & Feedback signal & Example feedback \\
\midrule
\texttt{click} 
& Target element, location, navigation or new-tab effect 
& \texttt{Succeed: `click` on <button> "Search" at (512, 86) executed. Page navigated to ...} \\

\texttt{click} 
& No-op detection 
& \texttt{Succeed: `click` on <div> at (211, 335) executed. Note: no visible navigation or new tab detected.} \\

\hline \\

\texttt{write} 
& Focused element and typed content 
& \texttt{Succeed: `write` typed "Alpine Ridge" into <input> role=combobox "Search".} \\

\texttt{write} 
& Actual value mismatch 
& \texttt{Note: the field's actual value is "New York, NY", which differs from the typed text.} \\

\hline \\

\texttt{scroll} 
& Scroll direction, amount, and boundary detection 
& \texttt{Succeed: `scroll` down by 50\% executed. Note: page scroll position did not change and may be at a boundary.} \\

\hline \\

\texttt{press\_keys} 
& Normalized keys and navigation effect 
& \texttt{Succeed: `press\_keys` [`Enter`] executed. Page navigated to ...} \\

\hline \\

\texttt{goto\_url} 
& Direct navigation success or failure 
& \texttt{Failed: `goto\_url` execution failed: Page.goto: net::ERR\_HTTP2\_PROTOCOL\_ERROR ...} \\

\hline \\

\texttt{tab ops} 
& New, switched, or closed tab state 
& \texttt{Succeed: `switch\_tab` switched to tab 1 (https://example.com).} \\

\hline \\

\texttt{done} 
& Task termination with final response 
& \texttt{Succeed: task marked as done. Response: ...} \\
\bottomrule
\end{tabular}
\label{tab:action-feedback}
\end{table}

\subsection{Textual Environment Feedback} \label{app:env_feedback}

Browser agents operate inside an instrumented web environment, which exposes state signals that are not directly visible from screenshots. These signals include the DOM element under an action coordinate, the currently focused input field, the active URL, tab state, and lightweight accessibility-tree changes. \textbf{If the agent relies only on visual input, it often needs multiple historical screenshots to infer whether the current state has changed after an action.} For example, after a \texttt{click}, a screenshot alone may not clearly indicate whether the intended element was selected, whether an input field was focused, or whether navigation was blocked. Similarly, after a \texttt{scroll}, the agent may not know whether the page actually moved or has already reached the boundary. Without such feedback, the agent can easily repeat the same failed click or continue scrolling in the current state. To provide this information more efficiently, we augment visual observations with lightweight textual environment feedback.

For each parsed tool call, the environment generates feedback by comparing browser states before and after action execution. Before executing an action, it records lightweight state variables such as the active page URL, the number of open tabs, the current scroll position, the focused element, and, when applicable, the DOM element located at the action coordinates. The environment then executes the corresponding Playwright operation and applies action-specific rule logic to convert the observed state change into a concise textual message. When an action contains multiple tool calls, the environment processes them sequentially and returns the feedback as a list, with each entry corresponding to one executed tool call.

Several representative feedback rules are summarized below and in Table~\ref{tab:action-feedback}. For \texttt{click}, the feedback uses the clicked coordinate, the DOM element returned by \texttt{document.elementFromPoint}, changes in the page URL, and changes in the number of open tabs. For \texttt{write}, the feedback reports the currently focused element and compares the intended text with the actual value in the field after typing. For \texttt{scroll}, the feedback compares pre- and post-action scroll offsets to identify successful scrolling, boundary conditions, or no-op actions. For \texttt{press\_keys}, the environment reports the normalized key sequence and whether the action triggers a URL change.

Execution exceptions are also caught and converted into explicit failure messages rather than silently terminating the rollout. Successful actions return concise messages describing both execution status and observable browser-state changes. The resulting feedback is appended to the next observation, giving the policy a compact signal for detecting failed clicks, blocked navigation, unintended tab changes, rejected inputs, boundary scrolling, and other common open-web interaction issues.

\subsection{Action Space}\label{app:action_space}
We define a compact browser action space consisting of 13 atomic tools, as
summarized in Table~\ref{tab:browser-tools}. These tools cover the core
operations needed for open-web interaction, including pointer control, keyboard
input, page navigation, tab management, and task termination. Each action is
represented as a structured tool call with explicit arguments, which makes the
policy output easy to parse and execute in the browser environment.

\begin{table}[h]
    \centering
    \caption{Browser action space: 13 atomic tools grouped by family.}
    \label{tab:browser-tools}
    \small
    \begin{tabular}{@{}p{0.14\columnwidth}p{0.11\columnwidth}p{0.17\columnwidth}p{0.5\columnwidth}@{}}
    \toprule
    \textbf{Category} & \textbf{Tool} & \textbf{Argument} & \textbf{Description} \\
    \midrule
    \multirow{3}{*}{Pointer Mgmt.}
                     & \texttt{click}        & \texttt{x, y, button, click\_type} & Mouse-click at a screen pixel; supports single/double click and left/right/middle button. \\
                     & \texttt{hover}        & \texttt{x, y} & Move the cursor to a pixel to reveal tooltips or open dropdowns. \\
                     & \texttt{drag}         & \texttt{x1, y1, x2, y2} & Drag-and-drop from a start pixel to an end pixel. \\
    \midrule
    \multirow{2}{*}{Keyboard Mgmt.}
                     & \texttt{write}        & \texttt{text} & Clear the focused input and type a string. \\
                     & \texttt{press\_keys}  & \texttt{keys} & Press keys sequentially or as a hotkey combo. \\
    \midrule
    \multirow{4}{*}{Page Nav.}
                     & \texttt{scroll}       & \texttt{direction, amount} & Scroll the page or element by a viewport fraction. \\
                     & \texttt{goto\_url}    & \texttt{url} & Navigate the current tab to a given URL. \\
                     & \texttt{go\_back}     & --- & Navigate back in the browser history. \\
                     & \texttt{wait}         & \texttt{seconds} & Pause for $N$ seconds to allow the page to settle. \\
    \midrule
    \multirow{3}{*}{Tab Mgmt.}
                     & \texttt{new\_tab}     & --- & Open a new blank browser tab. \\
                     & \texttt{switch\_tab}  & \texttt{index} & Switch to the tab with the given 0-based index. \\
                     & \texttt{close\_tab}   & --- & Close the current tab. \\
    \midrule
    Termination      & \texttt{done}         & \texttt{answer} & End the episode and emit the final answer. \\
    \bottomrule
    \end{tabular}
\end{table}

\subsection{Reward Design} \label{app:reward_design}
Our browser RL reward is a hybrid of deterministic rule-based verification and
LLM-as-a-judge evaluation. The deterministic rules serve two purposes: they
enforce the interaction protocol required by the browser environment, and they
avoid unnecessary judge calls for trajectories that are clearly invalid or
incomplete. \textbf{The judge model is only used for completed trajectories that pass
these basic validity checks.}
For each generated trajectory $\tau_i$, we compute three reward-related
quantities: a format score $F(\tau_i)$, a judge score $J(\tau_i)$, and the final
training reward $R(\tau_i)$. 

\paragraph{Rule-based format verification.}
Before invoking the judge model, we verify that the agent response follows the
required browser-action format. In our response
format, a turn is considered valid only if it contains the expected closing
thinking tag and a parseable tool call. This ensures that the model emits an
executable browser action rather than free-form text. 

\paragraph{Rule-based status filtering.}
The judge model is not called for trajectories that do not reach a completed
browser state, where agent outputs 'done' by itself. If the rollout terminates because of malformed tool calls,
generation aborts, context-length truncation, environment errors, or exhausting
the maximum number of browser steps, the judge score is set to zero by rule:
\[
J(\tau_i) = 0.
\]
Malformed format termination is treated as a special failure case and receives
a negative final reward:
\[
R(\tau_i) = -1
\quad
\text{if the trajectory terminates due to repeated format errors.}
\]
This penalizes trajectories that cannot produce executable browser actions.

Even for completed trajectories, the judge model is skipped if no final answer
can be extracted. The final answer must come
from the terminal \texttt{done} tool call. If the final answer is missing, we set
$J(\tau_i)=0$ without querying the judge.

\paragraph{VLM-as-a-judge evaluation.}
For completed trajectories with a valid final answer, we query a judge model to
determine task success. The judge input contains the original task instruction,
the agent's final answer, three recent screenshots from the trajectory, and an explicit action history summarizing the agent's
tool calls and environment feedback. The judge is instructed to return a verdict
of either \texttt{SUCCESS} or \texttt{NOT SUCCESS}.

The judge output is mapped back to a binary score by a deterministic parser:
\[
J(\tau_i) =
\begin{cases}
1, & \text{if the judge output contains \texttt{SUCCESS}},\\
0, & \text{if the judge output contains \texttt{NOT SUCCESS}},\\
0, & \text{if the judge output cannot be parsed}.
\end{cases}
\]
The parser checks for \texttt{NOT SUCCESS} before \texttt{SUCCESS} to avoid
mistaking a negative verdict for a positive one.

\paragraph{Final reward composition.}
The final reward is a gated combination of the rule-based checks and the judge
score:
\[
R(\tau_i) =
\begin{cases}
-1, & \text{if the trajectory terminates due to repeated format errors},\\
0, & \text{if } F(\tau_i)=0,\\
J(\tau_i), & \text{otherwise}.
\end{cases}
\]
Thus, a trajectory can receive reward $1$ only if it both follows the required
browser-action format and is judged successful.

\paragraph{Invalid or unreliable samples.}
If reward evaluation fails because of judge timeout or infrastructure-related
rollout failures, the sample can be marked for removal from optimization. Such
samples have their loss mask zeroed, so they do not contribute gradients. This
keeps reward-model or environment-system failures from introducing noisy policy
updates.

\begin{table}[h!]
\centering
\small
\setlength{\tabcolsep}{5pt}
\renewcommand{\arraystretch}{1.12}
\caption{Training hyperparameters for browser-agent reinforcement learning. Values summarize the Qwen3-VL 4B browser training configuration in \texttt{examples/browser}; ranges indicate script variants.}
\begin{tabular}{ll}
\toprule
\textbf{Hyperparameter} & \textbf{Value} \\
\midrule
Number of training iterations & 90 \\
Max rollout steps  & 15, 30 \\
Context screenshots per turn & 1 \\
Judge screenshots per trajectory & 3 \\
Judge model & GPT-4.1 or our trained 8B judge model \\
Effective rollout query number per iteration & 48 \\
Group size per prompt & 5 \\
PPO epochs & 2 \\
Maximum response length & 1024  \\
Maximum context length & 32768  \\
RL algorithm & MM-GRPO \\
KL coefficient & 0.0 \\
Entropy coefficient & 0.0 \\
PPO clipping range & low 0.2, high 0.28 \\
Global batch size for optimization & 256 \\
Optimizer & Adam \\
Learning rate & $1\times 10^{-6}$, constant schedule \\
Weight decay & 0.1 \\
Adam betas & $(0.9, 0.98)$ \\
Training backend & Megatron \\
Tensor parallel size & 4 \\
Micro-batch size & 1 \\
Attention backend & FlashAttention \\
Browser viewport & $1280\times1000$, DPR 1 \\
Coordinate normalization scale & 1000 \\
\hline
Sampling temperature & 0.8 \\
Rollout engine & SGLang \\
Sandbox mode & Kubernetes sandbox \\
Sandbox CPU and memory & 1 CPU, 4 GiB \\
Max concurrent sandboxes & 80--100 during training \\
Inference step timeout & 30 s \\
Rollout task timeout & 600 s \\
Sandbox acquire timeout & 600 s \\
Browser step request timeout & 45 s \\
Browser step retries & 2 \\
Environment exit timeout & 15 s \\
Rollout abort wait timeout & 30 s \\
Rollout health check & Interval 30 s, timeout 30 s, first wait 180 s \\
\bottomrule
\end{tabular}
\label{tab:browser_training_hyperparameters}
\end{table}

\begin{table}[h!]
\centering
\small
\setlength{\tabcolsep}{5pt}
\renewcommand{\arraystretch}{1.12}
\caption{Evaluation hyperparameters for browser-agent evaluation in \texttt{examples/browser}.}
\begin{tabular}{ll}
\toprule
\textbf{Hyperparameter} & \textbf{Value} \\
\midrule
Max evaluation rollout steps & 30 \\
Sandbox mode & Kubernetes sandbox \\
Sandbox CPU and memory & 1 CPU, 4 GiB \\
Parallel browser sandboxes & 16 \\
Context screenshots per turn & 1 \\
Judge screenshots per trajectory & 3 \\
Judge model & GPT-4.1 for our evaluation success rate w/o aborted tasks; \\  &  officially recommended models for  reporting the official score \\
Judge timeout & 120 s \\
Generation temperature & 0.6 for the official-score evaluation and 0.0 for evaluation success rate w/o aborted tasks \\
Top-$p$ & 0.95 for the official-score evaluation and 1.0 for evaluation success rate w/o aborted tasks \\
Top-$k$ & 20 for the official-score evaluation and 1 for evaluation success rate w/o aborted tasks \\
Maximum response length & 4096 tokens \\
Maximum context length & 32768 tokens \\
Task timeout & 600 s \\
Inference step timeout & 120 s \\
Rollout engine & SGLang \\
SGLang dtype & bfloat16 \\
SGLang context length & 32768 tokens \\
SGLang tensor parallel size & 2 \\
SGLang data parallel size & 4 \\
\bottomrule
\end{tabular}
\label{tab:browser_evaluation_hyperparameters}
\end{table}

\subsection{Training Details}\label{app:train_details}

\textbf{SFT.} We train Qwen3-VL-4B-Thinking for $3$ epochs with peak learning rate $10^{-5}$ under a cosine schedule with a $10\%$ linear warmup. 
Each optimizer step uses a per-device batch of $2$ with $8$-step gradient accumulation, giving a per-worker effective batch of $16$ and a global batch of 128 across 8 data-parallel workers. For the 8B model, we use the same training setup, with the only difference being the SFT dataset: we add $500$ trajectories from the InSTA-v3 subset, resulting in $912$ trajectories in total.

\textbf{MM-GRPO.} The policy is initialized from the SFT 
checkpoint. During training, each task is executed interactively in a live browser 
environment. Each rollout state comprises the current observation, recent screenshot 
context, and environment feedback in browser tool-call format. Rewards combine format 
validity with an LLM-as-a-judge success signal, where the judge receives the full 
action history and recent trajectory screenshots as input. To ensure stable training, 
we apply bounded rollout execution with explicit timeouts governing model generation, 
browser actions, sandbox acquisition, and task-level rollout completion. Training runs 
for 90 iterations and requires approximately 300 B200 GPU hours in total. We employ 
trajectory-level dynamic sampling, collecting rollouts until 48 effective groups of 
size 5 are obtained. Key training hyperparameters for our OpenWebRL-4B are summarized in 
Table~\ref{tab:browser_training_hyperparameters}. Regarding the 8B model, we use a lower learning rate of $5\times 10^{-7}$.

\subsection{Evaluation Details}\label{app:eval_details}

Each evaluation task is executed as an interactive browser trajectory with a maximum 
of 30 steps. We serve our model using the SGLang framework.
For \textbf{official success rate evaluation}, we follow the judge protocol of prior 
work and use Browser-Use Stealth Browsers. Based on the finding of~\cite{gupta2026molmoweb} 
that stochastic sampling outperforms deterministic sampling for web agents, we adopt 
stochastic decoding with the following hyperparameters: temperature 0.6, top-$p$ 0.95, 
top-$k$ 20, max response length 4096 tokens, and repetition penalty 1.0.
For \textbf{success rate evaluation without aborted tasks}, we use deterministic decoding with 
temperature 0.0.
All evaluation settings and decoding parameters are summarized in 
Table~\ref{tab:browser_evaluation_hyperparameters}.

\section{Benchmark Details} \label{app:benchmark_details}
We evaluate on three online benchmarks:
\begin{itemize}
    \item \textbf{WebVoyager}~\citep{he2024webvoyager} is a 643-task open-domain benchmark spanning 15 popular websites. We use the version curated by FARA~\citep{fara7b2025}, which removes infeasible tasks and updates those with outdated information from the original dataset, resulting in 595 tasks in total\footnote{\scriptsize\url{https://github.com/microsoft/fara/blob/44908264c810d3806365c6aab63a43c2d52a8057/webeval/data/webvoyager/WebVoyager\_data\_08312025.jsonl\#L4}}. For evaluation, we follow the canonical WebVoyager protocol\footnote{\scriptsize\url{https://github.com/MinorJerry/WebVoyager/blob/main/evaluation/auto\_eval.py}} and use GPT-4o as the judge, consistent with FARA.
    \item \textbf{Online-Mind2Web}~\citep{xue2025an} comprises 300 longer-horizon tasks drawn from 136 popular websites across diverse domains. Compared to WebVoyager, these tasks typically require more extended multi-step reasoning and interaction sequences. For evaluation, we adopt the OSU AgentTrek protocol\footnote{\scriptsize\url{https://github.com/OSU-NLP-Group/Online-Mind2Web/blob/main/src/methods/agenttrek\_eval.py}}, following the standard setup used in prior work with o4-mini as the judge.
    \item \textbf{DeepShop}~\citep{lyu2025deepshop} is a 150-task benchmark focused on online shopping scenarios, where agents must search, compare, and select products under realistic constraints. Task success is determined using Molmo-Web’s structured-output evaluation protocol~\citep{gupta2026molmoweb} with GPT-4o as the judge, which programmatically verifies the correctness of the final response\footnote{\scriptsize\url{https://github.com/allenai/molmoweb/blob/main/benchmarks/judges/deepshop_judge.py}}.
\end{itemize}


\section{Additional Experiment Results}\label{app:additional_exp}

\subsection{Task Difficulty Understanding}\label{app:task_difficulty}

\begin{figure}[h]
    \centering
    \begin{minipage}[c]{0.33\linewidth}
        \centering
        \includegraphics[width=\linewidth,keepaspectratio]{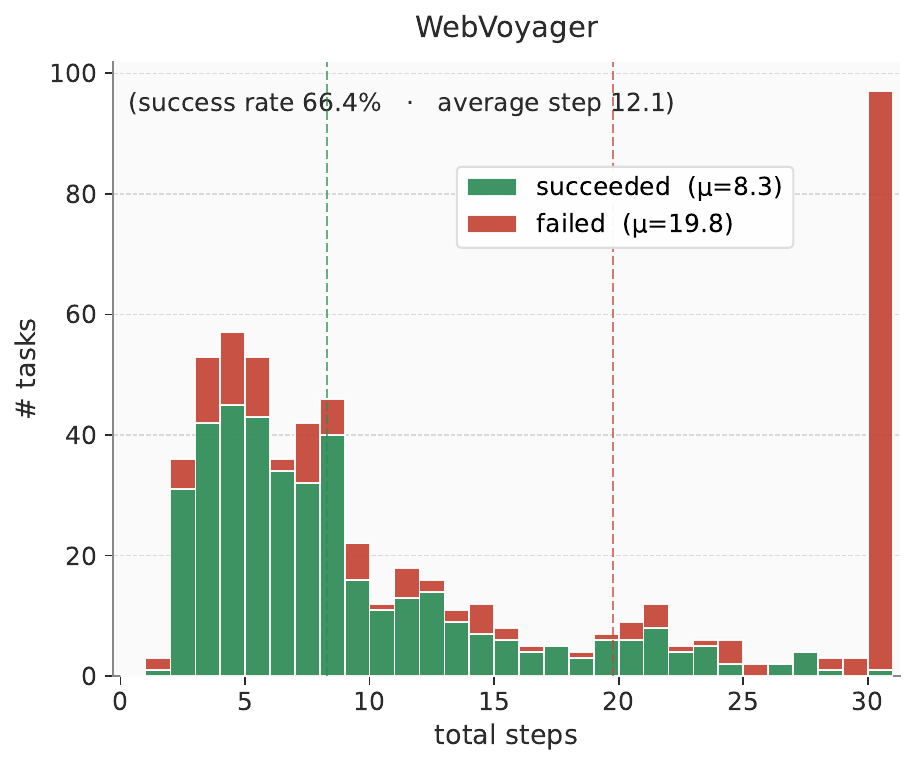}
    \end{minipage}\hfill
    \begin{minipage}[c]{0.33\linewidth}
        \centering
        \includegraphics[width=\linewidth,keepaspectratio]{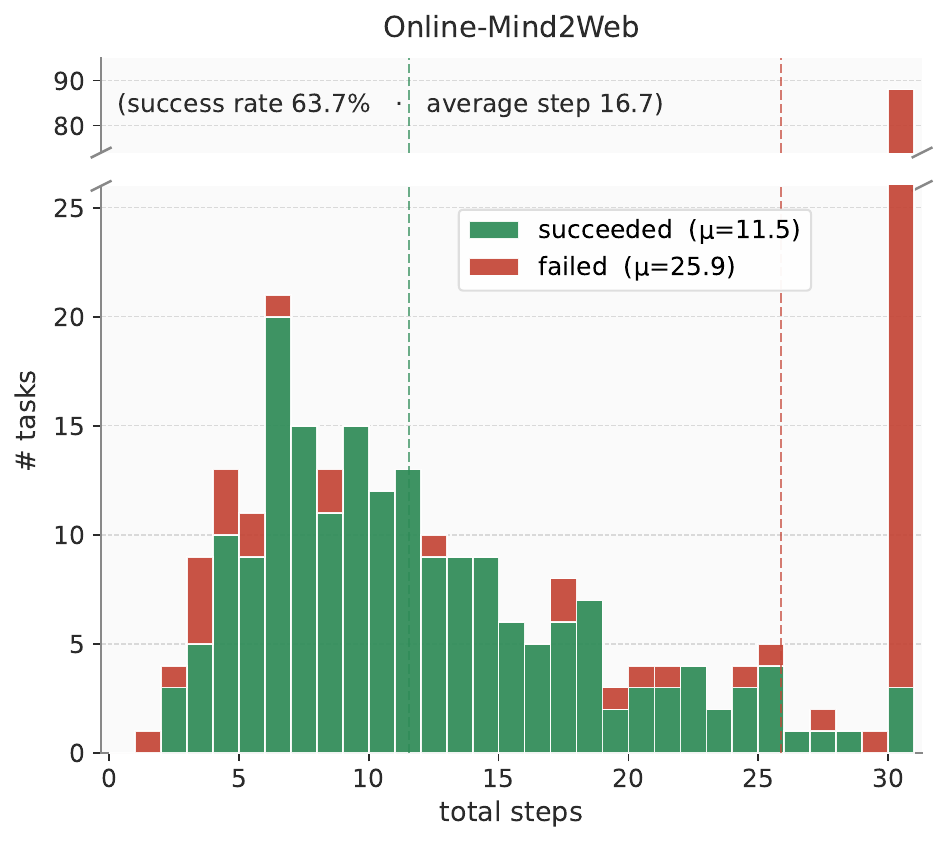}
    \end{minipage}\hfill
    \begin{minipage}[c]{0.33\linewidth}
        \centering
        \includegraphics[width=\linewidth,keepaspectratio]{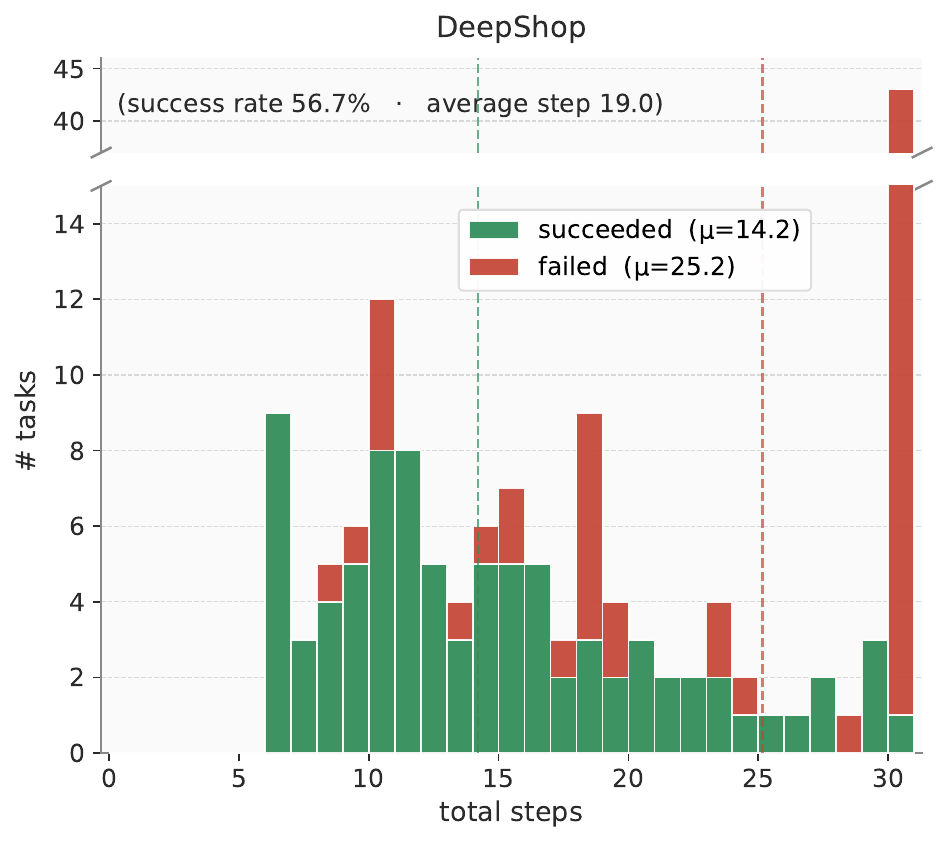}
    \end{minipage}
    \caption{Trajectory length (total steps) distribution of Qwen3-VL-235B-Thinking.}
    \label{fig:step-distribution}
\end{figure}

To characterize task difficulty and the number of steps required for task completion, we analyze Qwen3-VL-235B-Thinking's results across three online benchmarks. As shown in Figure~\ref{fig:step-distribution}, Online-Mind2Web and DeepShop exhibit lower success rates and require more steps to complete successfully (11.5 and 14.2 on average, respectively) compared to WebVoyager (8.0). Note that the step counts reported in Figure~\ref{fig:step-distribution} are averaged over all trajectories, including failed ones. These results confirm that Online-Mind2Web and DeepShop are more challenging benchmarks, demanding longer action sequences and yielding lower success rates. The stronger performance of OpenWebRL-4B on precisely these harder benchmarks (Table~\ref{tab:browser-results}) therefore provides stronger evidence for the effectiveness of our training framework.

\subsection{Results on AgentRewardBench}

To further evaluate the generalizability and effectiveness of our judge, we report results on AgentRewardBench~\cite{lu2025agentrewardbench}. AgentRewardBench contains 1,302 trajectories from five widely used web-agent benchmarks: WebArena (WA), VisualWebArena (VWA), AssistantBench (AB), WorkArena (Work), and WorkArena++ (Wk++). As shown in Table~\ref{tab:arb-results}, OpenWebRL-Judge-8B achieves the best overall F1 and Recall, while maintaining competitive Precision across benchmarks. OpenWebRL-Judge-4B also performs strongly, achieving the best Precision on WebArena and WorkArena.

It is worth noting that AgentRewardBench is built on a different agent stack, introducing distribution shift in both action representation and available environment feedback. In particular, AgentRewardBench contains action types outside our action space, making some action conversions approximate. Its trajectories also lack the per-step environment feedback used by our judges, leaving only partial signals such as URL changes, propagated errors, and accessibility-tree fragments. We believe these results should be viewed as a lower-bound estimate of our performance under this out-of-distribution setting.

\begin{table}[h!]
\centering
\caption{Fine-grained evaluation results on AgentRewardBench. We report overall Precision, Recall, and F1, as well as per-benchmark Precision across AssistantBench (AB), VisualWebArena (VWA), WebArena (WA), WorkArena (Work), and WorkArena++ (Wk++). Rows marked with $^{\dagger}$ are taken from AgentRewardBench~\cite{lu2025agentrewardbench} of its simple judge with final screenshot, and rows marked with $^{*}$ are taken from Online-Mind2Web~\cite{xue2025an}.}
\label{tab:arb-results}

\setlength{\tabcolsep}{4.5pt}
\renewcommand{\arraystretch}{1.12}

\begin{tabular}{lccc|ccccc}
\toprule
\multirow{2}{*}{Judge}
& \multicolumn{3}{c|}{Overall}
& \multicolumn{5}{c}{Per-Benchmark Precision} \\
\cmidrule(lr){2-4}
\cmidrule(lr){5-9}
& Precision & Recall & F1
& AB & VWA & WA & Work & Wk++ \\
\midrule

Claude 3.7 S.$^{\dagger}$
& 69.4 & 76.3 & 72.7
& 71.4 & 64.8 & 69.3 & 85.3 & 66.7 \\

GPT-4o$^{\dagger}$
& 68.1 & 80.3 & 73.7
& 77.8 & 60.7 & 69.9 & 93.8 & 59.6 \\

GPT-4o Mini$^{\dagger}$
& 64.5 & 78.3 & 70.8
& 80.0 & 57.4 & 66.9 & 90.3 & 54.8 \\

WebJudge (GPT-4o)$^{*}$
& 73.7 & 71.2 & 72.4
& 66.7 & 69.8 & 72.6 & 92.3 & 75.0 \\

WebJudge-7B$^{*}$
& 75.7 & 58.0 & 65.6
& 80.0 & 66.7 & 77.5 & 100.0 & 70.0 \\

\textbf{OpenWebRL-Judge-4B}
& 73.2 & 67.0 & 70.0
& 75.0 & 62.2 & 78.0 & 92.6 & 70.0 \\

\textbf{OpenWebRL-Judge-8B}
& 72.8 & 72.5 & 72.7
& 80.0 & 67.0 & 73.5 & 82.9 & 74.2 \\

\bottomrule
\end{tabular}
\end{table}

\subsection{Comparison with Online Filtered Behavior Cloning Baseline}

\begin{figure}
    \centering
    \includegraphics[width=0.65\linewidth]{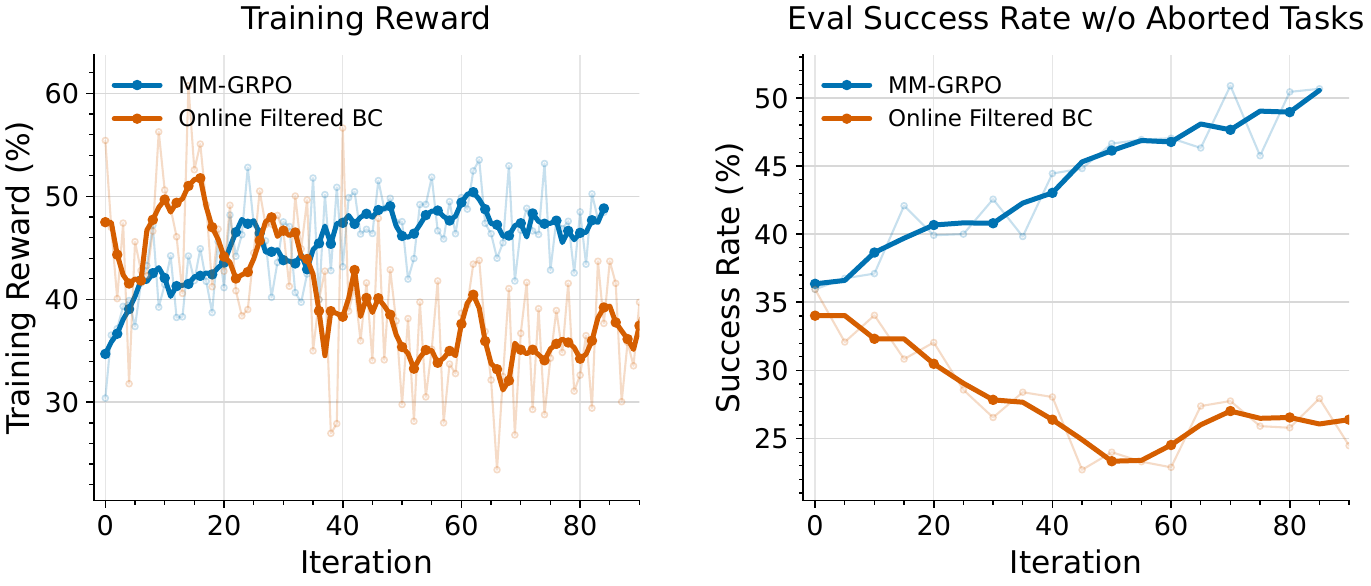}
    \caption{Comparison with online filtered behavior cloning baseline.}
    \label{fig:rejectsamplingbaseline}
\end{figure}

Figure~\ref{fig:rejectsamplingbaseline} compares MM-GRPO with an online filtered behavior cloning (BC) baseline. The BC baseline uses the same online training setup as MM-GRPO on top of the same SFT checkpoint, but only keeps successful trajectories for supervised updates, uses a fixed rollout query batch size, and does not apply dynamic sampling. Although online filtered BC initially achieves competitive training rewards, its evaluation success rate without aborted tasks steadily declines throughout training, eventually dropping below 30\%. In contrast, MM-GRPO maintains stable optimization and continuously improves evaluation performance, reaching over 50\% evaluation success rate, indicating that simply imitating filtered online trajectories is insufficient for robust policy improvement in open-web environments. These results highlight the importance of reinforcement learning updates, rather than pure supervised imitation, for effective online adaptation and long-horizon web interaction.

\section{Proxy-Based Analysis of Response Length}\label{app:proxy_analysis}
To better understand why response length increases across training iterations,
we define four lightweight lexical proxies over agent responses. These proxies
are not intended to be exhaustive semantic labels; instead, they provide
interpretable, automatically computable indicators of common reasoning patterns
observed in long responses. We apply them to assistant turns by matching
precision-oriented regular expression families in the reasoning text before the
first tool call. These proxies are deliberately simple and transparent. A
response may match multiple proxies, and the categories are not mutually
exclusive. For example, a response can both summarize failed attempts and
diagnose a CAPTCHA blocker. We therefore use these proxies as descriptive
measurements rather than ground-truth semantic labels. We summarize those
proxies in Table~\ref{tab:length-proxies}.

\begin{table}[t]
\centering
\small
\caption{Lexical proxies used to analyze response-length growth. The proxies are regex-based, interpretable, and not mutually exclusive.}
\begin{tabular}{p{0.18\linewidth} p{0.26\linewidth} p{0.27\linewidth} p{0.21\linewidth}}
\toprule
\textbf{Proxy} & \textbf{What it measures} & \textbf{Example phrase families} & \textbf{Motivation} \\
\midrule
History summarization &
The agent lists previous attempts, actions, failures, or exhausted sources. &
``I've tried'', ``we already checked'', ``previous attempt'', ``so far'', ``multiple approaches'', ``all sources failed''. &
Long responses often contain detailed inventories of prior failures, especially near termination. \\

Blocker diagnosis &
The agent diagnoses CAPTCHA, verification, access denial, or automation-related blockers. &
``CAPTCHA'', ``Cloudflare'', ``403 forbidden'', ``verification required'', ``anti-bot'', ``automated access'', ``cannot solve verification''. &
Many long-tail failures arise from web security or anti-automation barriers. \\

Retry-plan loop &
The agent considers repeated alternatives, direct navigation, or ``one more'' attempts. &
``one final approach'', ``try a different search'', ``alternative source'', ``go back'', ``search directly'', ``let me reconsider''. &
Captures cases where the agent continues planning retries instead of terminating or taking a concise action. \\

Condition proof &
The agent checks task constraints one by one or explicitly verifies that concrete constraints are satisfied. &
``meets the requirements'', ``all criteria met'', ``first condition'', ``under \$50'', ``within 5 miles'', ``gluten-free''. &
Verbose successful or partially successful cases often explicitly verify each user constraint. \\
\bottomrule
\end{tabular}
\label{tab:length-proxies}
\end{table}

All regular expressions are evaluated
case-insensitively. A step is marked as proxy-bearing if it matches any pattern
from one of the four proxy families. We also separately track responses with
none of the four proxies as a baseline. This no-proxy baseline shows smaller
length growth than proxy-bearing responses, suggesting that these four patterns
account for a meaningful portion of the observed per-step verbosity increase.

In our analysis, the proxies already appear in early iterations, indicating
that \textbf{later training does not introduce entirely new reasoning modes.
Instead, later iterations tend to amplify these patterns when they occur.} In
particular, history summarization and retry-plan loops are associated with
longer responses and heavier response-length tails. Blocker diagnosis
captures common failure states caused by CAPTCHA or access restrictions, while
condition proof explains why even valid or successful responses may become more
verbose: the agent increasingly justifies its answer by checking task
requirements explicitly.

\paragraph{Exact matching patterns.}
For reproducibility, we include the exact patched proxy definitions used in the
analysis below. We use a compact style to keep long regular expressions readable.

\lstdefinestyle{proxypython}{
  language=Python,
  basicstyle=\ttfamily\scriptsize,
  columns=fullflexible,
  keepspaces=true,
  breaklines=true,
  breakatwhitespace=false,
  showstringspaces=false,
  frame=single,
  framerule=0.3pt,
  xleftmargin=0.5em,
  xrightmargin=0.5em,
  aboveskip=0.6em,
  belowskip=0.6em,
  captionpos=b
}



\begin{lstlisting}[style=proxypython,caption={History summarization proxy patterns.}]
HISTORY_SUMMARIZATION = [
    r"\b(?:i(?:'ve| have)|we(?:'ve| have))\s+(?:already\s+)?"
    r"(?:tried|attempted|checked|searched|looked at|visited|used)\b",
    r"\b(?:i|we)\s+already\s+"
    r"(?:tried|attempted|checked|searched|looked at|visited|used)\b",
    r"\b(?:i|we)\s+had\s+(?:already\s+)?"
    r"(?:tried|attempted|checked|searched|looked at|visited|used)\b",
    r"\b(?:previous|earlier|last)\s+"
    r"(?:attempt|search|approach|step|site|page)\b",
    r"\b(?:so far|up to this point),?\s+(?:i|we)\s+(?:have|'ve|had)\b",
    r"\b(?:tried|attempted|checked|searched|looked at|visited|used)\b"
    r"[^.!?\n]{0,80}\b(?:multiple|several)\s+"
    r"(?:attempts|approaches|searches|sites|websites|sources)\b",
    r"\b(?:multiple|several)\s+"
    r"(?:attempts|approaches|searches|sites|websites|sources)\b"
    r"[^.!?\n]{0,80}\b(?:failed|blocked|led to|resulted in|did not|"
    r"didn't|couldn't|cannot)\b",
    r"\b(?:all|every)\s+(?:attempt|approach|site|source|path|option)s?\s+"
    r"(?:has|have)\s+(?:failed|led to|resulted in)\b",
    r"\b(?:exhausted|tried)\s+(?:all|every|multiple|several)\s+"
    r"(?:options|approaches|paths|sources|sites)\b",
]
\end{lstlisting}

\begin{lstlisting}[style=proxypython,caption={Blocker diagnosis proxy patterns.}]
BLOCKER_DIAGNOSIS = [
    r"\b(?:captcha|recaptcha|cloudflare|access denied|403\s+forbidden|"
    r"robot check)\b",
    r"\baccess\s+(?:is\s+)?forbidden\b",
    r"\b(?:blocked|blocking|unable to access|cannot access|can't access|"
    r"not accessible)\b",
    r"\b(?:(?:security|bot|human)\s+verification|verification\s+"
    r"(?:challenge|required|page|screen|check))\b",
    r"\b(?:security check|security measure|anti[- ]bot|anti[- ]automation)\b",
    r"\b(?:automated access|automated behavior|bot detection|"
    r"detected automated)\b",
    r"\b(?:requires?|needs?)\s+human\s+"
    r"(?:interaction|verification|input)\b",
    r"\b(?:cannot|can't|unable to)\s+(?:interact with|solve|complete)\s+"
    r"(?:the\s+)?(?:captcha|recaptcha|verification)\b",
]
\end{lstlisting}

\begin{lstlisting}[style=proxypython,caption={Retry-plan loop proxy patterns.}]
RETRY_PLAN_LOOP = [
    r"\b(?:one|a)\s+(?:final|last)\s+(?:approach|attempt|try|option)\b",
    r"\b(?:try|attempt|use|take)\s+(?:a\s+)?"
    r"(?:different|another|alternative|new)\s+"
    r"(?:approach|search|query|site|source|path|method)\b",
    r"\b(?:need|needs|needed|should|could|might|may|will|would|try|use|"
    r"take|switch to|look for|search for)\b[^.!?\n]{0,50}\b"
    r"(?:different|alternative|another|new)\s+"
    r"(?:approach|search|query|site|source|path|method)\b",
    r"\b(?:different|alternative|another|new)\s+"
    r"(?:approach|search|query|site|source|path|method)\b"
    r"[^.!?\n]{0,50}\b(?:is needed|may work|might work|could work|"
    r"should work|would help|to try|to search|to use)\b",
    r"\b(?:go back|return)\s+(?:and|to)\s+(?:try|search|look|navigate)\b",
    r"\b(?:navigate directly|search directly|try searching directly|"
    r"search more broadly)\b",
    r"\b(?:let me|i should)\s+"
    r"(?:reconsider|try again|try another|switch to|use a different)\b",
    r"\b(?:rather than|instead of)\s+[^.!?\n]{0,80}\b"
    r"(?:i should|i'll|i will|let me)\s+(?:try|search|navigate|use)\b",
]
\end{lstlisting}

\begin{lstlisting}[style=proxypython,caption={Condition proof proxy patterns and composite rules.}]
PROOF_VERB = (
    r"\b(?:meets?|matches?|satisf(?:y|ies|ied)|fulfills?|"
    r"qualif(?:y|ies|ied)|confirms?|verif(?:y|ies|ied)|fits?|"
    r"complies with|addresses)\b"
)
REQ_WORD = (
    r"(?:criteria|requirements|conditions|constraints|criterion|"
    r"requirement|condition|constraint)"
)
REQ_MET = PROOF_VERB + r"[^.!?\n]{0,80}\b" + REQ_WORD + r"\b"
ALL_MET = (
    r"\b(?:all|each|every)\s+(?:the\s+)?"
    r"(?:criteria|requirements|conditions|constraints)\b"
    r"[^.!?\n]{0,80}\b(?:met|satisfied|fulfilled|addressed|checked)\b"
)
ORDINAL_REQ = (
    r"\b(?:(?:first|second|third|fourth|fifth|sixth|seventh|eighth|"
    r"ninth|tenth|final|last)|\d+(?:st|nd|rd|th)?)\s+\b"
    + REQ_WORD + r"\b"
)

VALUE_PATTERNS = [
    r"\b(?:under|less than|below)\s+\$?[0-9][0-9,]*(?:\.[0-9]+)?\b",
    r"\b(?:at least|at most|no more than|fewer than|greater than)\s+"
    r"[0-9][0-9,]*(?:\.[0-9]+)?\b",
    r"\bwithin\s+[0-9][0-9,]*(?:\.[0-9]+)?\s*"
    r"(?:miles?|mi|km|kilometers?)\b",
    r"\b(?:before|after)\s+[0-9]{1,2}:[0-9]{2}\b",
    r"\b(?:model years?|years?)\s+[0-9]{4}\s*(?:-|to)\s*[0-9]{4}\b",
]
DOMAIN_PATTERNS = [
    r"\bgood with kids\b",
    r"\bgood with cats\b",
    r"\bgluten[- ]free\b",
    r"\bnut[- ]free\b",
    r"\bshort[- ]haired\b",
    r"\bdepart(?:ing|s)? from\s+[A-Z][A-Za-z .-]+",
    r"\blocated near\s+[A-Z0-9][A-Za-z0-9 .-]+",
]

def has_condition_proof(step):
    if re.search(REQ_MET, step, flags=re.I):
        return True
    if re.search(ALL_MET, step, flags=re.I):
        return True
    if re.search(ORDINAL_REQ, step, flags=re.I):
        return True

    constraints = VALUE_PATTERNS + DOMAIN_PATTERNS
    for sent in re.split(r"[.!?\n]+", step):
        has_proof = re.search(PROOF_VERB, sent, flags=re.I)
        n_constraints = sum(
            bool(re.search(pat, sent, flags=re.I)) for pat in constraints
        )
        has_value = any(
            re.search(pat, sent, flags=re.I) for pat in VALUE_PATTERNS
        )
        if has_proof and has_value:
            return True
        if has_proof and n_constraints >= 2:
            return True
    return False
\end{lstlisting}

\begin{lstlisting}[style=proxypython,caption={Final proxy and non-proxy labels.}]
def proxy_labels(step):
    labels = set()
    if any(re.search(p, step, flags=re.I) for p in HISTORY_SUMMARIZATION):
        labels.add("history_summarization")
    if any(re.search(p, step, flags=re.I) for p in BLOCKER_DIAGNOSIS):
        labels.add("blocker_diagnosis")
    if any(re.search(p, step, flags=re.I) for p in RETRY_PLAN_LOOP):
        labels.add("retry_plan_loop")
    if has_condition_proof(step):
        labels.add("condition_proof")
    return labels

def is_non_proxy(step):
    return len(proxy_labels(step)) == 0
\end{lstlisting}

\paragraph{Limitations.}
Because the proxies are regex-based, they may miss paraphrases or include mild
false positives. The retry-plan proxy is intentionally broad enough to recover
context-anchored alternative-approach language, but this also means that some
benign planning statements may be counted. The condition-proof proxy is more
conservative: generic mentions of constraints are not sufficient unless they
match explicit requirement-satisfaction patterns or co-occur with concrete
constraints and a proof verb in the same sentence. Nevertheless, the proxy
definitions are interpretable, easy to reproduce, and useful for identifying
high-level mechanisms behind response-length growth.

\section{Error Analysis}
\label{app:error-analysis}

\begin{figure}[h]
    \centering
    \includegraphics[width=0.55\linewidth]{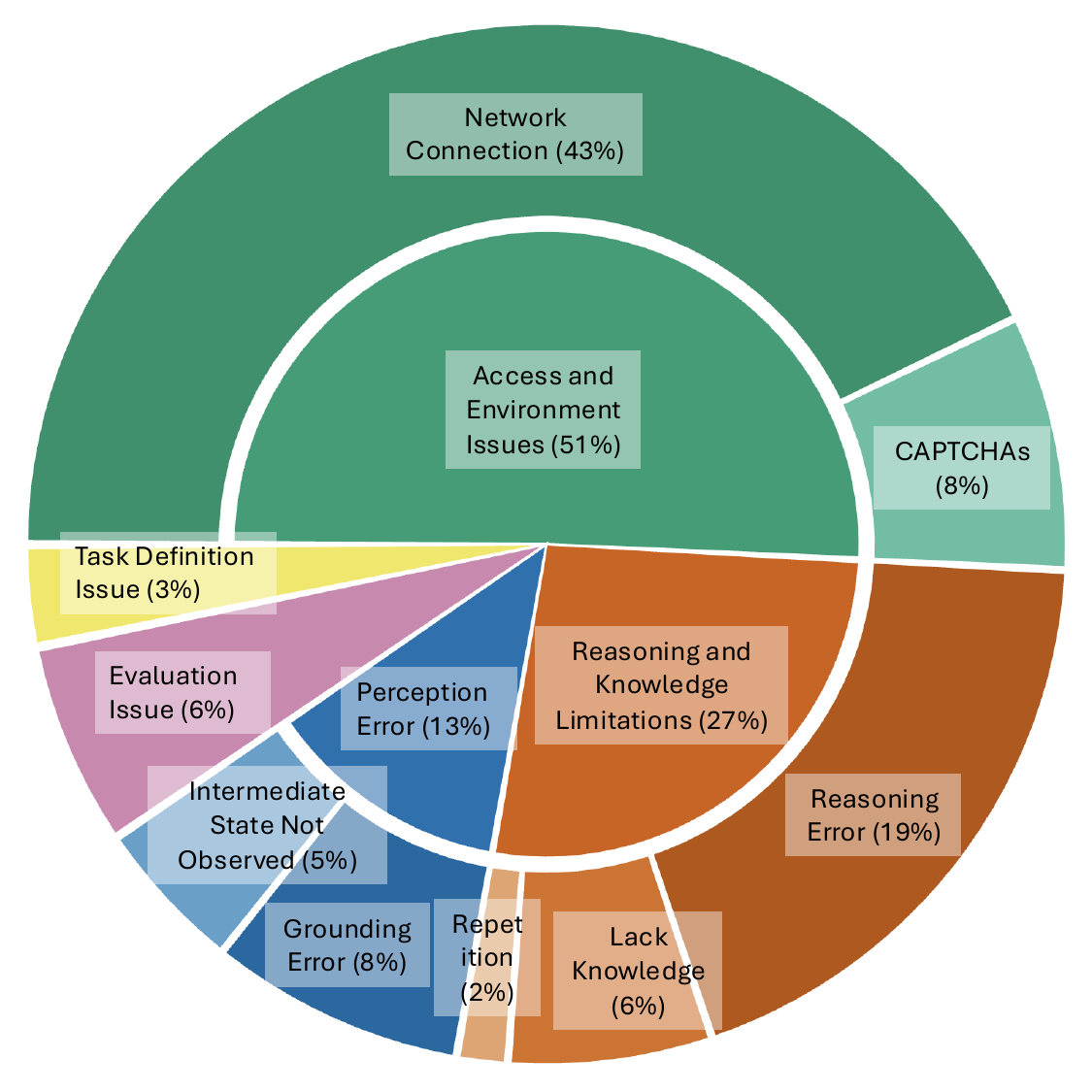}
    \caption{Distribution of failure modes based on a manual inspection of 100 failed trajectories.}
    \label{fig:failure-case-analysis}
\end{figure}

We manually inspect 100 failed trajectories sampled from evaluation runs of multiple OpenWebRL checkpoints on Online-Mind2Web without the Browser-Use Stealth Browser service, and categorize them into four failure types. Figure~\ref{fig:failure-case-analysis} summarizes the distribution.

\textbf{Access and environment issues (51\%).}
The largest category stems from live-web instability, including page loading failures, 
access restrictions, and CAPTCHA blockers. In these cases, the agent typically follows 
a reasonable strategy, but environmental obstacles prevent the trajectory from reaching 
a verifiable terminal state.

\textbf{Reasoning and knowledge limitations (27\%).}
The second major category reflects model-side deficiencies in planning and constraint 
tracking. In long-horizon tasks---particularly shopping or search tasks involving 
multiple simultaneous requirements such as price, color, rating, size, and product 
type---the agent may satisfy most constraints while overlooking one or more. A smaller 
share of failures arises when the task presupposes background or domain-specific 
knowledge that the model does not reliably possess.

\textbf{Visual grounding and interaction errors (13\%).}
Another common failure mode is inaccurate interaction with the page. The agent may click a nearby but incorrect element, miss a small dropdown or pagination control, or fail to notice that a filter has not been applied. These errors are more likely when multiple actions are executed before the next observation, making it harder for the agent to diagnose intermediate failures.

\textbf{Task definition and evaluation issues (9\%).} The remaining failures stem from  ambiguous or underspecified task instructions, with a small fraction attributed to judge errors---cases where the agent completes the task correctly but the automated judge incorrectly marks it as incomplete.

Overall, this error analysis indicates that further progress requires greater robustness to 
open-web instability, more reliable long-horizon constraint maintenance, improved 
intermediate-state monitoring, and stronger recovery mechanisms for failed interactions.

\newpage

\section{Prompt Templates}

\noindent This section presents prompt examples used in our pipeline. For rollout examples, we show one user turn and the corresponding assistant turn in the exact serialized conversation format consumed by the model. For judge examples, we provide the full template together with one concrete example constructed from a real trajectory.

\subsection{Agent Input and Output}\label{app:agent_input_output}

\begin{promptbox}{Example Rollout Input: System Prompt}
{\ttfamily\small
\noindent\textless |im\_start|\textgreater system\\
You are a GUI agent designed to operate in an iterative loop to automate browser tasks.\\
\\
\# GUI Agent Policy\\
\\
As an autonomous GUI agent operating on the Web Browser platform, your primary
function is to analyze screen captures and perform appropriate UI actions to
complete assigned tasks.\\
\\
\#\# Core Responsibilities\\
\\
You can perform web browser interactions including mouse interactions,
keyboard interactions
, navigation, tab management, task completion, and
waiting.\\
\\
\#\# Input Information\\
\\
At each step, you will receive the following information:\\
1. Action History\\
2. User Request\\
3. Observation, including tab info, screenshot, and optional A11y Tree\\
\\
\#\# Output Requirements\\
\\
Your output must include one \textless think\textgreater\ block and one or
more \textless tool\_call\textgreater\ blocks. The \\
response must follow this
exact structure:\\
\textless think\textgreater\ ... \textless /think\textgreater\\
\textless tool\_call\textgreater\ \{"name": ..., "arguments": ...\}
\textless /tool\_call\textgreater\\
\\
\#\# Guidelines\\
\\
Analyze the current browser state, reflect on prior actions, assess progress
toward the goal
, plan the next step, and emit only valid executable tool
calls. When appropriate, multiple sequential tool calls may be emitted in one
response. The final \texttt{done.response} must be plain text only, with no
LaTeX-style markup.\\
\\
\# Tools\\
...
\\
\textless |im\_end|\textgreater
}
\end{promptbox}

\begin{promptbox}{Example Rollout Input: User Prompt}
{\ttfamily\small
\noindent\textless |im\_start|\textgreater user\\
\textless tool\_response\textgreater\\
Succeed: \textasciigrave click\textasciigrave\ on \textless rect\textgreater\ at (237, 310) executed. Note: no visible navigation or new\\ tab detected.\\
Succeed: \textasciigrave click\textasciigrave\ on \textless button\textgreater\ type=button "Apply" at (393, 557) executed. Page navigated to https://www.healthgrades.com/usearch.\\
\\
\textless observation\textgreater\\
screen size: 1280 x 1000\\
Current tab: 0\\
Available tabs:\\
- Tab 0 (active): https://www.healthgrades.com - Healthgrades \textgreater\ Universal Search \textgreater\ Find a Doctor\\
screenshot:\\
\textless |vision\_start|\textgreater\textless |image\_pad|\textgreater\textless |vision\_end|\textgreater\\
\textless /observation\textgreater\\
\textless /tool\_response\textgreater\textless |im\_end|\textgreater
}
\vspace{4pt}
\begin{center}
\promptboximage[0.5\linewidth]{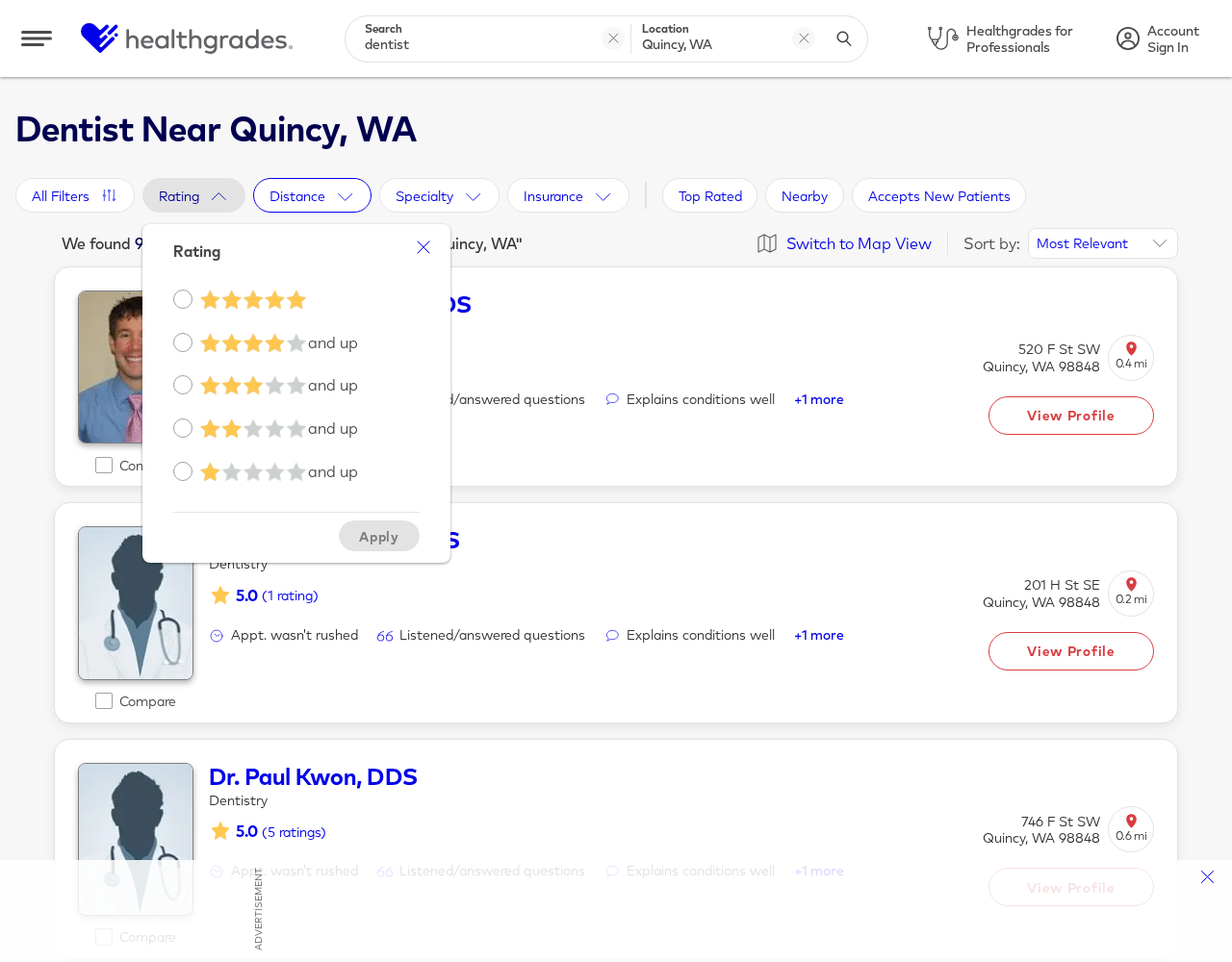}{Screenshot attached in the raw rollout input above}
\end{center}
\end{promptbox}

\begin{promptbox}{Example Rollout Output}
{\ttfamily\small
\noindent\textless |im\_start|\textgreater assistant\\
\textless think\textgreater\\
The search has successfully found dentists with 5-star ratings near Quincy,
WA (zip code \\98011 area). There are exactly 2 results shown, both with 5.0
ratings. The first dentist, Dr. Jay Magnuson, is located at a distance of
0.2 miles, while the second, Dr. Paul Kwon, is 0.6 miles away.\\
\\
The user asked for the closest 5-star rated dentist to zip code 98011.
Based on the search results, Dr. Jay Magnuson is the closest at 0.2 miles.
This completes the task of finding the closest 5-star rated dentist.\\
\\
I should now provide the final answer with the name and location of the
closest 5-star rated dentist.\\
\\
\textless /think\textgreater\\
\textless tool\_call\textgreater\\
\{"name": "done", "arguments": \{"response": "The closest 5-star rated dentist to zip code 98011 is Dr. Jay Magnuson, DDS, located at 201 H St SE, Quincy, WA 98848, which is 0.2\\ miles away. He has a 5.0 rating based on 1 review."\}\}\\
\textless /tool\_call\textgreater\textless |im\_end|\textgreater
}
\end{promptbox}

\subsection{OpenWebRL-Judge Input and Output}\label{app:judge_input_output}

\begin{promptbox}{Prompt Template for OpenWebRL-Judge}
\noindent As an evaluator, you will be presented with four primary components to assist you in your role:

\begin{enumerate}
    \item \textbf{Web Task Instruction}: This is a clear and specific directive provided in natural language, detailing the online activity to be carried out.
    \item \textbf{Agent Action History}: This is a chronological summary of the agent's observed actions across steps. Use it to understand what the agent tried to do, but do not treat it as ground truth if it conflicts with the screenshots.
    \item \textbf{Result Screenshots}: This is a visual representation of the screen showing the result or intermediate state of performing a web task. Each screenshot will be annotated with an inferred step index in text.
    \item \textbf{Result Response}: This is a textual response obtained after the execution of the web task.
\end{enumerate}

\vspace{1pt}
\noindent \textbf{\#\#\# TASK: \{\}} \\
\noindent \textbf{\#\#\# Agent Action History:} \{\} \\
\noindent \textbf{\#\#\# Result Response: \{\}} \\
\noindent \textbf{\#\#\# \{num\} screenshots from the trajectory are attached below with inferred step indices.}
\vspace{1pt}

\noindent \textbf{\#\# Guidelines}
\begin{itemize}
    \item You DO NOT NEED to interact with web pages or perform actions.
    \item You SHOULD use the screenshots as the strongest evidence about the actual page state.
    \item You SHOULD use the action history to judge whether the agent followed the instruction and whether the final response is supported by what happened on screen.
    \item If the action history conflicts with screenshots, trust the screenshots.
    \item NOTE that the instruction may involve more than one task. Failing to complete either task should be considered unsuccessful.
    \item NOTE that the final response may contradict the screenshots; in that case the screenshots prevail. If the final response contains information not visible in the screenshots, you may still consider it only if it is consistent with the screenshots and action history.
    \item You should first explain your reasoning with explicit reference to the instruction, action history, screenshots, and final response. Then provide a definitive verdict as either \texttt{SUCCESS} or \texttt{NOT SUCCESS}.
\end{itemize}

\noindent \textbf{\#\# Response Format:} \\
Return the reasoning first, then the verdict.

\noindent \texttt{Reasoning:} \\
Explain how you arrived at your final evaluation with explicit reference to the instruction, action history, screenshots, and final response.

\noindent \texttt{Verdict:} \\
\texttt{SUCCESS} or \texttt{NOT SUCCESS} \\

\noindent \textbf{Your verdict:}

\end{promptbox}

\begin{promptbox}{OpenWebRL-Judge Example Input}
\noindent As an evaluator, you will be presented with four primary components to assist you in your role.

\noindent \textbf{\#\#\# TASK:} Find the closest 5-star rated dentist to zip code 98011. \\

\noindent \textbf{\#\#\# Agent Action History:} \\
Step 1: action=\texttt{[click(\{"point\_2d": [307, 630]\}); write(\{"message": "dentist"\});\\press\_keys(\{"keys": ["enter"]\})]} environment feedback: click: Succeed: \texttt{click} on search input executed; write: typed \texttt{"dentist"}; \texttt{press\_keys} navigated to Healthgrades search results for dentists in Quincy, WA. \\
Step 2: action=\texttt{[click(\{"point\_2d": [150, 207]\})]} environment feedback: clicked the \texttt{Rating} filter with no navigation. \\
Step 3: action=\texttt{[click(\{"point\_2d": [185, 310]\}); click(\{"point\_2d": [307, 557]\})]} environment feedback: selected the 5-star filter and clicked \texttt{Apply}, navigating to filtered 5-star dentist results in Quincy, WA. \\
Step 4: action=\texttt{[done(\{"response": "The closest 5-star rated dentist to zip code\\ 98011 is Dr. Jay Magnuson, DDS, located at 201 H St SE, Quincy, WA 98848, which\\ is 0.2 miles away. He has a 5.0 rating based on 1 review."\})]} environment feedback: task marked as done with the final response above. \\

\noindent \textbf{\#\#\# Result Response:} The closest 5-star rated dentist to zip code 98011 is Dr. Jay Magnuson, DDS, located at 201 H St SE, Quincy, WA 98848, which is 0.2 miles away. He has a 5.0 rating based on 1 review. \\

\noindent \textbf{\#\#\# 3 screenshots from the trajectory are attached below with inferred step indices.} \\

\vspace{4pt}
\begin{center}
\promptboximage{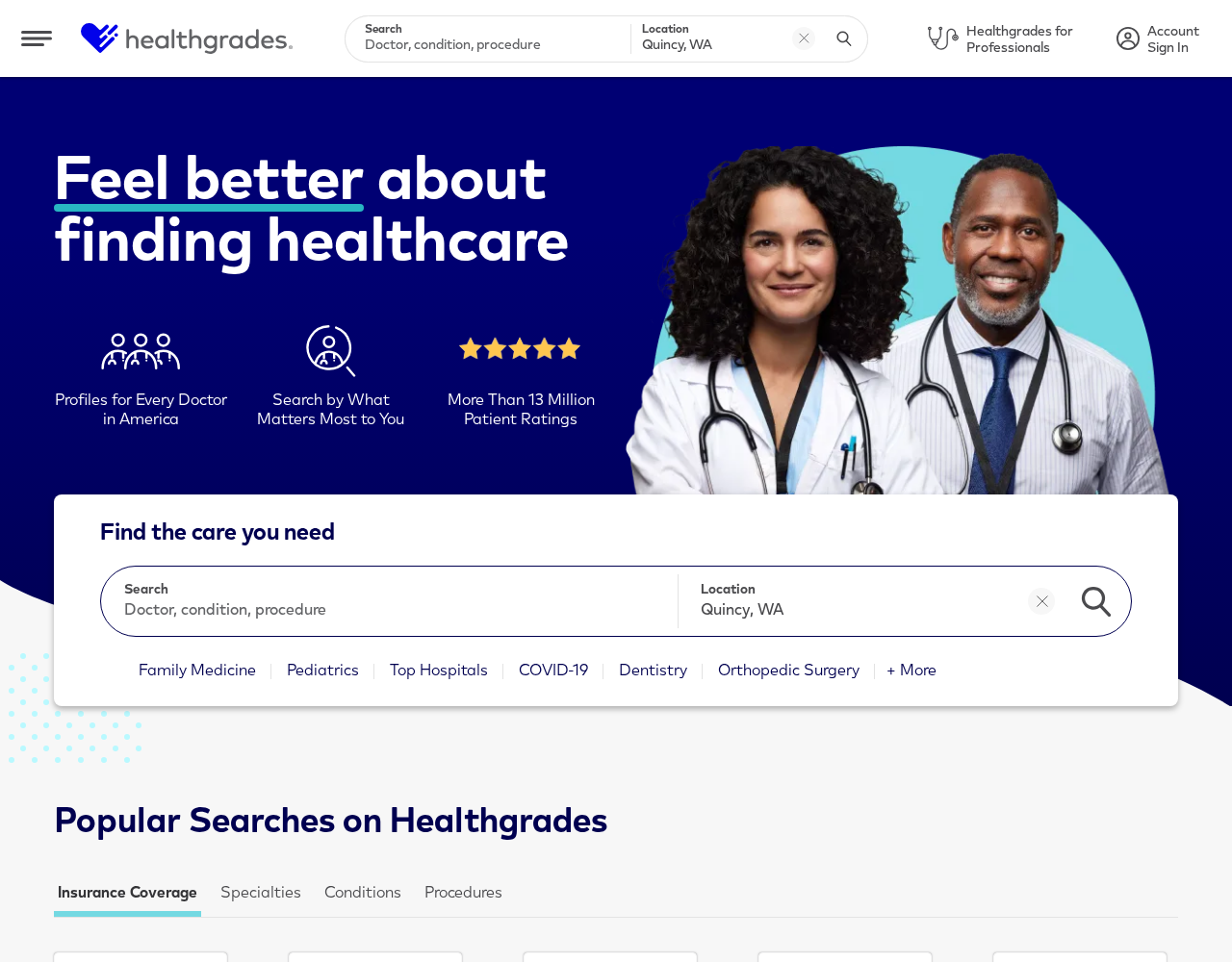}{Screenshot 1}
\hfill
\promptboximage{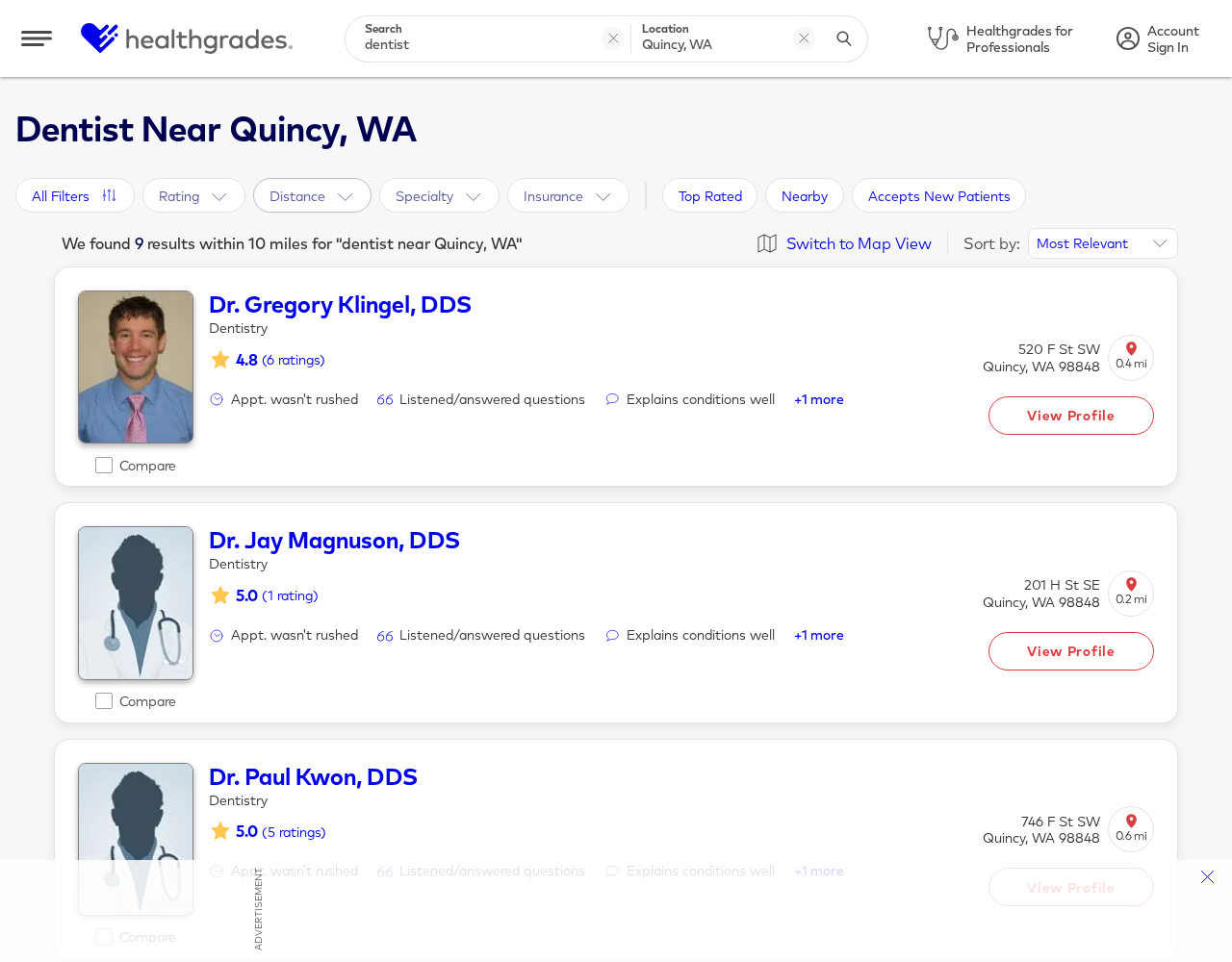}{Screenshot 2}
\hfill
\promptboximage{images/judge_example_dentist_step_3.png}{Screenshot 3}
\end{center}

\noindent \textbf{\#\# Guidelines}
... \\

\noindent \textbf{\#\# Response Format}
... \\

\noindent \textbf{Your verdict:}
\end{promptbox}

\begin{promptbox}{OpenWebRL-Judge Example Output}
\noindent \textbf{Reasoning:} The task instruction is to find the closest 5-star rated dentist to zip code 98011. The action history shows that the agent searched for \texttt{dentist}, opened the \texttt{Rating} filter, selected 5 stars, and returned Dr. Jay Magnuson, DDS as the answer. However, the screenshots show that the search location remained Quincy, WA rather than zip code 98011. Screenshot 2 shows dentists near Quincy, WA after the initial search, and Screenshot 3 confirms that the final 5-star results are also in Quincy, WA, with Dr. Jay Magnuson at 0.2 miles and Dr. Paul Kwon at 0.6 miles from that Quincy search center. The final response repeats Quincy, WA and does not provide evidence that the result is near zip code 98011. Since the instruction explicitly requires a dentist near 98011 and the observed trajectory instead stays in Quincy, WA, the task was not completed successfully.

\noindent \textbf{Verdict:} \texttt{NOT SUCCESS}
\end{promptbox}

\section{Broader Impacts}
OpenWebRL releases an open framework, training data, and a 4B-parameter model for training visual web agents with online multi-turn RL. We discuss the main societal considerations below.

\paragraph{Positive impacts.} The strongest visual web agents today are proprietary, with training recipes and weights closed. By releasing our full pipeline, training tasks, and OpenWebRL-4B, we lower the barrier for academic and independent researchers to study and build on online RL for visual agents. Showing that competitive performance is achievable with modest initialization data also reduces reliance on expensive human demonstrations, making this line of research more accessible to resource-constrained groups. Capable open web agents further serve as building blocks for assistive applications that help users who struggle with conventional web interfaces.

\paragraph{Potential negative impacts.} More capable visual web agents lower the technical barrier for automated scraping in violation of site terms of service, circumvention of bot-detection systems, and large-scale generation of inauthentic accounts or engagement. Multi-turn agents can also compound errors over long horizons, which is particularly concerning when deployed in settings involving purchases, form submissions, or other state-changing actions. Online RL on live websites additionally places traffic load on third-party servers.

\paragraph{Mitigations.} We trained on benchmark environments and task distributions designed for agent research, and rate-limited interactions during data collection. OpenWebRL-4B is released at a scale where capabilities remain bounded relative to frontier proprietary systems, limiting the marginal uplift our release provides to a sophisticated misuse actor. We provide model and dataset cards documenting intended research use and encourage downstream users to respect site terms of service and applicable laws, and to apply human-in-the-loop confirmation or domain restrictions when deploying agents in user-facing settings. We believe open release with clear documentation better supports community work on agent safety, bot-detection defenses, and misuse evaluation than a closed release would.

\section{Additional Case Studies}

\begin{figure}[h]
    \centering
    \includegraphics[width=0.9\linewidth]{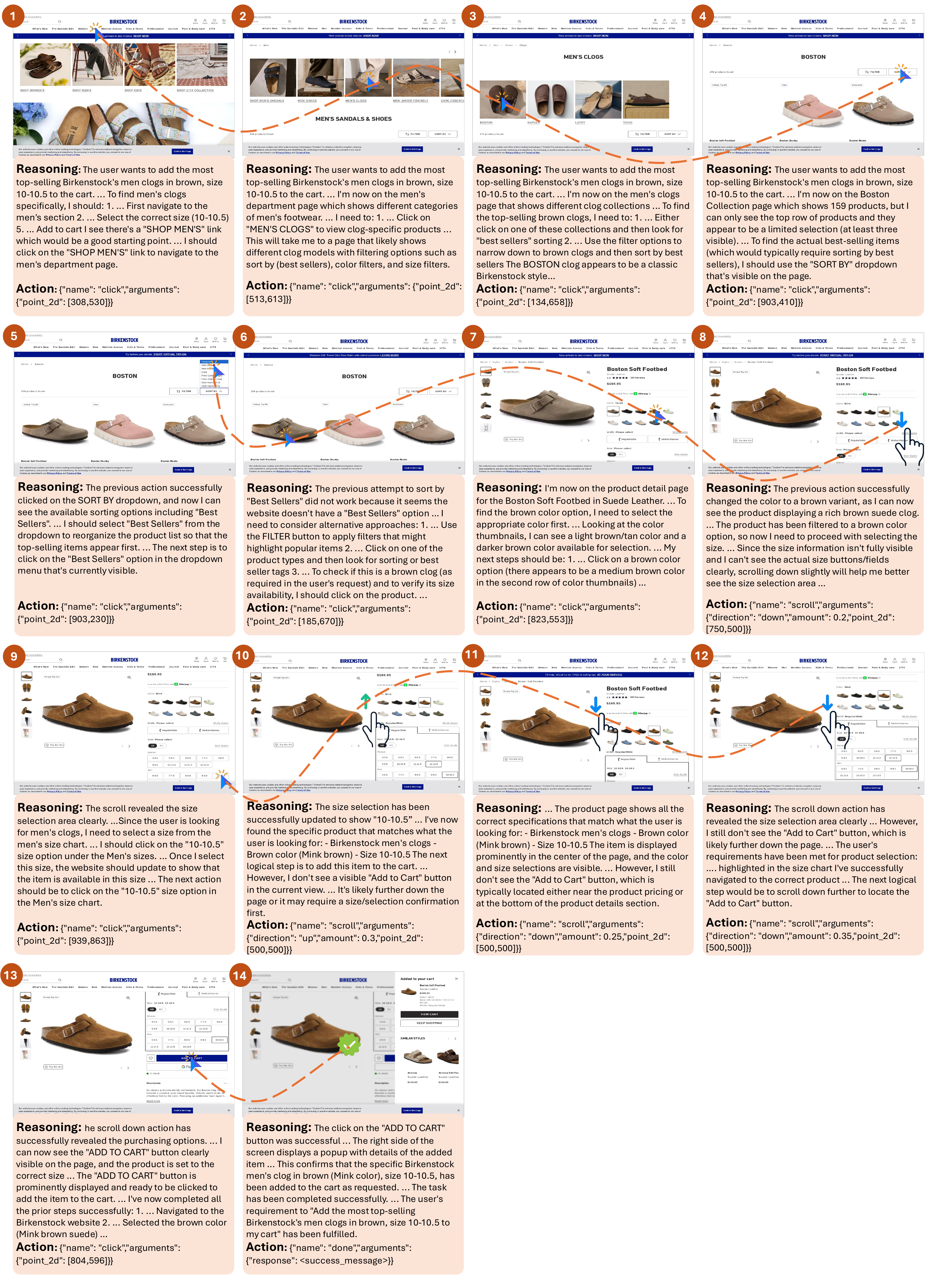}
    \caption{Trajectory Example: Add the most top-selling Birkenstock's men clogs in brown, size 10-10.5 to my cart.}
    \label{fig:case-study-clog}
\end{figure}

\begin{figure}
    \centering
    \includegraphics[width=0.9\linewidth]{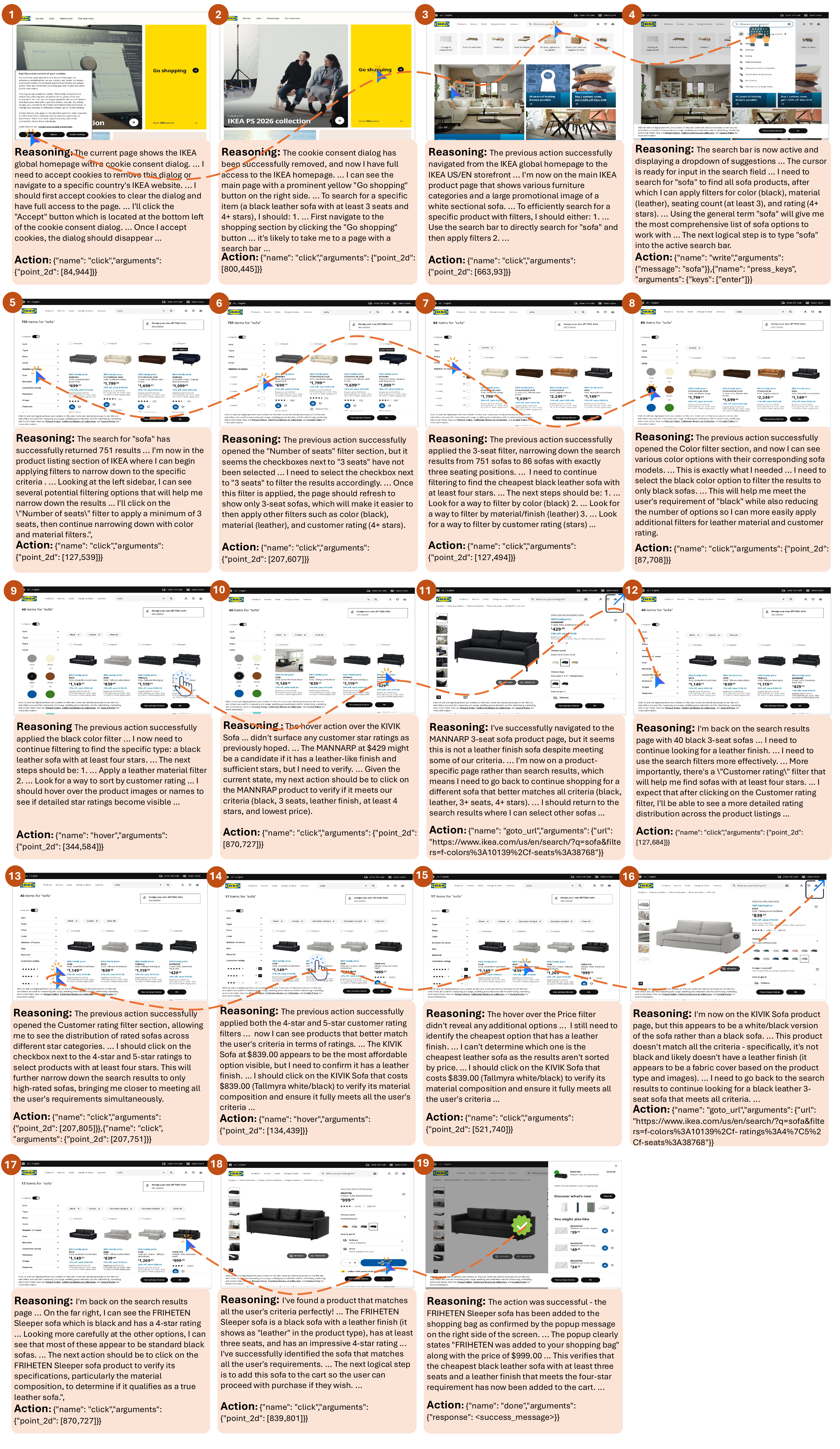}
    \caption{Trajectory Example: Add the cheapest black sofa with at least three seats, a leather finish, and at least four stars to my cart.}
    \label{fig:case-study-sofa}
\end{figure}

Figure~\ref{fig:case-study-clog} and Figure~\ref{fig:case-study-sofa} show two representative long-horizon shopping tasks in which the agent must satisfy multiple user constraints through extended web interaction. In Figure~\ref{fig:case-study-clog}, the agent searches for a top-selling men's Birkenstock clog in brown and size 10--10.5. In Figure~\ref{fig:case-study-sofa}, it searches for the cheapest black leather sofa with at least three seats and a rating of at least four stars. 

These trajectories highlight several agent capabilities. First, the agent decomposes the high-level request into concrete subgoals, including navigating to the relevant product category, applying filters, inspecting candidate products, selecting product options, and locating the final purchase control. Second, the agent exhibits reflective behavior by monitoring whether an action achieves the intended state and revising its strategy when it does not; for example, it switches from ineffective filtering or sorting operations to direct product inspection, and returns to the results page after identifying an unsuitable candidate. Third, the agent maintains task-relevant information across many steps, preserving constraints such as color, size, material, rating, seat count, and price while evaluating later pages. This persistent state tracking allows the agent to reject partially matching products and complete the task only after verifying that all required attributes are satisfied.

\end{document}